\documentclass[journal]{IEEEtran}
\IEEEoverridecommandlockouts

\usepackage{bbm}
\usepackage{amsmath}
\usepackage{amsfonts} 
\usepackage{amssymb}  
\usepackage{enumitem}
\usepackage{graphicx}
\usepackage{caption}
\usepackage{subcaption}
\usepackage{cite}

\def\BibTeX{{\rm B\kern-.05em{\sc i\kern-.025em b}\kern-.08em
    T\kern-.1667em\lower.7ex\hbox{E}\kern-.125emX}}

\begin{document}

\title{Elevator, Escalator, or Neither? Classifying Conveyor State Using Smartphone under \\
Arbitrary Pedestrian Behavior}

\author{Tianlang He,
        Zhiqiu Xia,
        S.-H. Gary Chan,~\IEEEmembership{Senior Member,~IEEE}
        \IEEEcompsocitemizethanks{
        \IEEEcompsocthanksitem This work was supported, in part, by Research Grants Council Collaborative Research Fund (under grant number C1045-23G). 
        \IEEEcompsocthanksitem Tianlang He and S.-H. Gary Chan are with The Hong Kong University of Science and Technology, Clear Water Bay, Kowloon, Hong Kong (e-mail: theaf@cse.ust.hk; gchan@cse.ust.hk).
        \IEEEcompsocthanksitem Zhiqiu Xia is with Rutgers University, New Brunswick, NJ, USA (e-mail: zx283@scarletmail.rutgers.edu).
        }}

\newcommand*{\sysname}{ELESON}

\maketitle

\begin{abstract}\label{sec: abstract}
Knowing a pedestrian's {\em conveyor state} of ``elevator,'' ``escalator,'' or ``neither'' is fundamental to many applications such as indoor navigation and people flow management.  
Previous studies on classifying the conveyor state often rely on specially designed body-worn sensors or make strong assumptions on pedestrian behaviors, which greatly strangles their deployability.  
To overcome this, we study the classification problem under arbitrary pedestrian behaviors using the inertial navigation system (INS) of the commonly available smartphones (including accelerometer, gyroscope, and magnetometer). 
This problem is challenging, because the INS signals of the conveyor states are entangled by the arbitrary and diverse pedestrian behaviors.
We propose \sysname{}, a novel and lightweight deep-learning approach that uses phone INS to classify a pedestrian to {\bf el}evator, {\bf es}calator, {\bf o}r {\bf n}either. 
Using causal decomposition and adversarial learning, \sysname{} extracts the motion and magnetic features of conveyor state independent of pedestrian behavior, based on which it estimates the state confidence by means of an evidential classifier.
We curate a large and diverse dataset with 36,420 instances of pedestrians randomly taking elevators and escalators under arbitrary unknown behaviors. Our extensive experiments show that \sysname{} is robust against pedestrian behavior, achieving a high accuracy of over 0.9 in F1 score, strong confidence discriminability of 0.81 in AUROC (Area Under the Receiver Operating Characteristics), and low computational and memory requirements fit for common smartphone deployment. 

\end{abstract}

\textbf{Keywords:} Conveyor state classification, smartphone, user behavior, IMU, magnetic field, causal representation learning, evidential model

\section{Introduction}\label{sec: intro}

Knowing whether a pedestrian is taking an elevator, escalator, or neither is fundamental to many smart city applications.  For example, in indoor navigation, such information enhances localization accuracy owing to better detection of floor transition~\cite{wang2024learning, zhang2024rloc, zhuo_fis_one_2023, zhuo2024online}.
Such knowledge also plays an important role in understanding pedestrian flow and conveyor preference in a venue, shedding insights on user journey, people management measures, and conveyor capacity planning~\cite{cai2023forecasting, wang2024coverage, he2022self}. 
However, previous studies on the subjects often employ specially designed body-worn sensors or make strong assumptions on pedestrian behavior, which greatly limits their wide applicability~\cite{kronenwett2018elevator, lang2018classifying, abdelnasser2015semanticslam, liu_smartcare_2015}. 

To overcome this, we study classifying a pedestrian into one of the three {\em conveyor states} of ``elevator,'' ``escalator,'' and ``neither'' without any behavior assumption, using the inertial navigation system (INS) commonly available from the off-the-shelf smartphone nowadays. Specifically, we use the multimodal INS readings from the accelerator (namely acceleration), gyroscope (namely angular velocity), and magnetometer (namely magnetic field) to classify the states.\footnote{We forgo barometer due to its relatively lower phone penetration and greater device heterogeneity, leaving it for future study~\cite{ye2014sbc, ye2018himeter}. }Note that we primarily focus on the conveyor states of elevator and escalator in indoor environments; readers interested in transportation modes (such as bus and flight travels) may refer to~\cite{zhang2021toward, li2021transportation, stenneth2011transportation} and the references therein. 

Conveyor state classification is challenging, because the measured INS readings are the mixture, or entanglement, of signals due to the two independent processes of conveyor state and arbitrary {\em pedestrian behavior}. 
In other words, the underlying process of conveyor state is continuously perturbed by various random and diverse behaviors of pedestrians, including, but not limited to, their spatial movements (such as walking, turning, and accommodating), phone carriage (whether held in hand, stored in pocket, or placed in bag), and various actions (such browsing, swinging, and shaking). These behaviors complicate and perturb the brittle conveyor signals, thereby obscuring the classification decision on conveyor states. 

Much effort has been made on using the smartphone INS to classify behaviors regarding human gesture recognition, gait detection, and action recognition~\cite{zhang2024mp, lu2019robust, andrade2022human, xu2023practically}.   
While impressive, their research problems are orthogonal to ours because a pedestrian's conveyor state is determined by the conveyor rather than his/her behaviors.\footnote{For example, while people could browse phones on elevators, the browsing behavior cannot be used to define the “elevator” state.} 
Moreover, these works treat the INS signal as a unit or aggregate in both the training and inference processes.  
If such a methodology is straightforwardly applied to our case, the accuracy would be unsatisfactory due to the perturbation caused by pedestrian behaviors (as confirmed in our extensive experimental results). 
Although the general approaches on classification robustness have been applied to INS processing, 
once extended to our problem, they require precise labels of behaviors during training~\cite{lu_out--distribution_2023, yang2022deep, chen2023domain, miao2024spatial, kaya2024human}.  
This is difficult to implement because exhaustively labeling the plethora of all the possible conceivable behaviors is prohibitively costly, next to impossible. Therefore, classifying the conveyor state under {\em arbitrary} pedestrian behaviors remains an open and challenging problem.  

\begin{figure}[tp]
    \centering
    \includegraphics[width=0.48\textwidth]{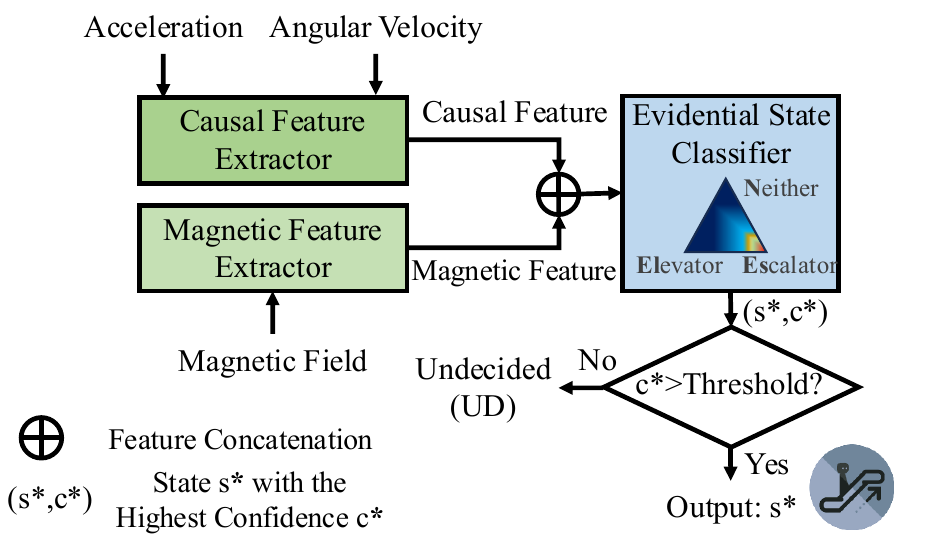}
    \caption{System overview of \sysname{}. }  
    \label{fig: system_diagram}
\end{figure}

We propose \sysname{}, a novel and lightweight deep-learning approach to classify a pedestrian of arbitrary behaviors to  {\bf el}evator, {\bf es}calator, {\bf o}r {\bf n}either using the multimodal INS readings from smartphone. 
We overview \sysname{} in Figure~\ref{fig: system_diagram}, which consists of three major modules:
\begin{enumerate}[leftmargin=*]
         \item \emph{Causal feature extractor to segregate the conveyor motion  from pedestrian behaviors:}
        The motion signals in terms of acceleration and angular velocity capture the movement of elevators and escalators.  
        However, as mentioned before, the motion feature is entangled with pedestrian behaviors.  
        Employing causal decomposition, we propose a causal feature extractor which segregates in deep feature space the motions of moving elevators and escalators, resulting in the causal feature.  We design a novel loss function so that such causal feature is extracted independent of pedestrian behaviors.

    \item \emph{Magnetic feature extractor to extract conveyor magnetic feature robust against pedestrian behaviors: } 
           The magnetic feature inside the metallic enclosed space of an elevator is different from that of a semi-open escalator. 
           However, some pedestrian behaviors (such as shaking or rotating of phone) often disrupt such magnetic feature.
           Using adversarial learning, we propose a magnetic feature extractor to capture the conveyor magnetic features robust against these behaviors. Extracting features from the temporal differential of the magnetic field signals, our extractor achieves generalizability for the elevators and escalators unseen in the training data. 
           
    \item \emph{Evidential state classifier to estimate the confidence of each state based on the causal and magnetic features:  }
            Applying evidence theory, we employ an evidential state classifier to 
            estimate the confidence of each conveyor state (between 0 and 1) given the causal and magnetic features.
            The pedestrian is classified to the state with the highest confidence (i.e., $s^*$ and $c^*$ in Figure~\ref{fig: system_diagram}) if it is above a certain threshold, or ``undecided''~(UD) otherwise. 
            In contrast to the conventional Softmax-based approaches, our classifier estimates the state confidence that reflects the similarity of the target INS signals with the training data, hence better discriminating the misclassifications. 
            
\end{enumerate}

We curate a large and diverse dataset with 36,420 instances from pedestrians randomly taking elevators and escalators with arbitrary behaviors. 
We conduct extensive experiments on the dataset, and show that
\sysname{} achieves high accuracy of over 0.9 in F1 score 
with a strong confidence discriminability of 0.81 in AUROC (Area Under the Receiver Operating Characteristics).  As compared with the state-of-the-art classification approaches (all treating the INS signals as a single unit),
  \sysname{} outperforms
significantly with~14\% improvement in F1 score.
We have also implemented  \sysname{} on mobile phones and demonstrated that it runs locally in real time with low computational overhead, requiring only 9MB of memory and consuming less than 2\% battery for a 2.5-hour operation. 

The remainder of this paper is organized as follows. 
We first review related work in Section~\ref{sec: relate}. 
Then, we present the problem, the causal feature extractor, and magnetic feature extractor in Section~\ref{sec: extractor}.  After that, we discuss the evidential state classifier to estimate state confidence in Section~~\ref{sec: detector}. 
Finally, we validate \sysname{} design with extensive experimental results in Section~\ref{sec: exp}, and
conclude in Section~\ref{sec: conclude}.

\section{Related Work}\label{sec: relate}

Previous studies on classifying the conveyor states often employ specially designed body-worn INS sensors or have specific strong assumptions on pedestrian  behaviors~\cite{lang2018classifying, liu_smartcare_2015}. For example, works in~\cite{lang2018classifying, kronenwett2018elevator} study the classification using foot-mounted sensors; works in~\cite{abdelnasser2015semanticslam, liu_smartcare_2015} are based on restricted user behaviors. However, these restrictions could limit their wide deployability. 
Recently, deep learning has shown powerful capabilities in INS signal processing, effectively classifying various pre-defined behaviors of phone users, including their actions, gestures, and gaits~\cite{zhang2021fine, prasanth2021wearable, chen2020subject, hong2024crosshar, khatun2022deep, haresamudram2021contrastive, saleem2023toward, demrozi2020human}. 
While impressive, these approaches for human behaviors cannot be satisfactorily extended to the conveyor states of elevator and escalator, because they consider the INS readings as a unit instead of a mixed signal. In our problem, the fragile signals of the underlying conveyor process are frequently perturbed by various arbitrary pedestrian behaviors, which makes it challenging to classify the states robustly.  
Furthermore, although domain generalization has been applied for robust INS classification, once extended to our problem, it requires precise and exhaustive labeling of pedestrian behaviors in the training process, which is, if not impossible, prohibitively difficult and costly~\cite{bento2022comparing, shen2023probabilistic, hu2023swl, lu_out--distribution_2023, yang2022deep, chen2023domain, miao2024spatial, kaya2024human}. 
Therefore, we propose \sysname{}, the first approach to address arbitrary pedestrian behaviors for conveyor state classification using phone INS without any need for behavior labeling.

Much research work has considered evidential classification for image classification, speech recognition, LiDAR/infrared object detection, etc.~\cite{bao2021evidential, wang2022uncertainty, pandharipande2023sensing, sun2022drone, xu2023ai}.  
Despite so, the evidential model for INS sensing has rarely been studied. In this paper, we use an evidential state classifier based on the causal and magnetic features of conveyor states and present a loss function for confidence estimation and sound classification.

\section{Causal and Magnetic Feature Extraction
}\label{sec: extractor}

In this section, we present the feature extraction of conveyor states under arbitrary pedestrian behaviors. 
After defining the problem in Section~\ref{subsec: def}, we discuss the causal feature extractor based on the acceleration and angular velocity in Section~\ref{subsec: motion}, and the magnetic feature extractor based on magnetic field in Section~\ref{subsec: magnetic}.

\subsection{Problem Definition}\label{subsec: def}

A phone-based inertial navigation system (INS) samples the acceleration, angular velocity, and magnetic field in the three dimensions at a fixed interval typically ranging from 1 to 20{\em ms}.
Given a sequence of the signals with $T$ time steps from a pedestrian's phone, or simply an {\em INS signal}, denoted as $x\in \mathbb{R}^{T\times 9}$, our overarching goal is to classify the {\em conveyor state} of the pedestrian, denoted as $s$. Specifically, we define $s$ as a categorical variable that can take one of the three values representing the states of ``elevator,'' ``escalator,'' or ``neither.''

To achieve this goal, a common practice is training a deep learning classifier that maps an INS signal to the conveyor state given a labeled dataset denoted as $D=\{(x_n, \tilde{s}_n)\}$, where $\tilde{s}_n$ is the label of the $n$th signal, or simply conveyor state label. After the training process, the classifier is considered ready for testing in real-world scenarios.  
By doing so, the underlying assumption is that the INS signals used for the training and testing are {\em independent and identically distributed} (IID), expressed as
\begin{equation}
    P(x_{test}\mid s)= P(x_{train}\mid s), 
\end{equation}
where $x_{test}$ and $x_{train}$ refer to the INS signals in testing and training scenarios, respectively. 

However, the IID assumption may be violated in conveyor state classification due to the {\em pedestrian behavior}. Specifically, the diverse and arbitrary behaviors of pedestrians often lead to a discrepancy of INS signals in the training and testing, which violates the IID assumption.  
Formally, under the impact of the conveyor state and pedestrian behavior, the conditional probability of obtaining an INS signal ($x$) is presented as
\begin{equation}
    P\left(x\mid s, V_p\right),       
\end{equation}
where $V_p$ is the variable of pedestrian behavior, and we consider its value drawn from an {\em uncountable set}.\footnote{This is because the impact of pedestrian's various behaviors on INS signal (such as their spatial movements, phone carriage styles, actions, and user heterogeneity) is difficult to {\em precisely} enumerate. } Though in the same conveyor state ($s$), it is hard to guarantee the pedestrian behaviors ($V_p$) in testing to match those in training, thus leading to the signal discrepancy, i.e. $P(x_{test}\mid s)\neq P(x_{train}\mid s)$. Such a discrepancy violates the IID assumption, which makes the deep learning classifier unreliable.  

To tackle the discrepancy, we apply a feature extraction module before classification.  
The module aims to extract a {\em conveyor state feature}, denoted as $z$, that is statistically independent of pedestrian behavior, shown as
\begin{equation}
    P\left( z\mid s, V_p\right) = P(z\mid s).    
\end{equation}
The signal discrepancy hence can be bridged by the extracted features in that
\begin{equation}
    P(z_{test}\mid s) = P\left( z_{train}\mid s \right), 
\end{equation}
where $z_{test}$ and $z_{train}$ are the conveyor state features extracted from testing and training scenarios, respectively. 

The module has two feature extractors, as shown in Figure~\ref{fig: system_diagram}. 
First, we divide an input INS signal into the motion signal (i.e., acceleration and angular velocity), denoted as $x_m\in \mathbb{R}^{T\times 6}$, and magnetic field signal, denoted as $x_b\in \mathbb{R}^{T\times 3}$, shown as
\begin{equation}
    x=[x_m, x_b],
\end{equation}
where $[\cdot,\cdot]$ is the concatenation operation. 
Then, a causal feature extractor captures a {\em conveyor causal feature}, denoted as $z_c$, from the motion signal, 
and a magnetic feature extractor captures a {\em conveyor magnetic feature}, denoted as $z_b$, from the magnetic field signal.   
Finally, the two features are concatenated to be a conveyor state feature, shown as
\begin{equation}
    z=[z_c, z_b].   
\end{equation}
After the feature extraction, we input the conveyor state feature to the evidential state classifier to be discussed in Section~\ref{sec: detector}. 
Next, we present the two feature extractors in detail.  

\subsection{Causal Feature Extractor}\label{subsec: motion}
\subsubsection{Overview}

\begin{figure}[tp]
    \centering
        \includegraphics[width=0.35\textwidth]{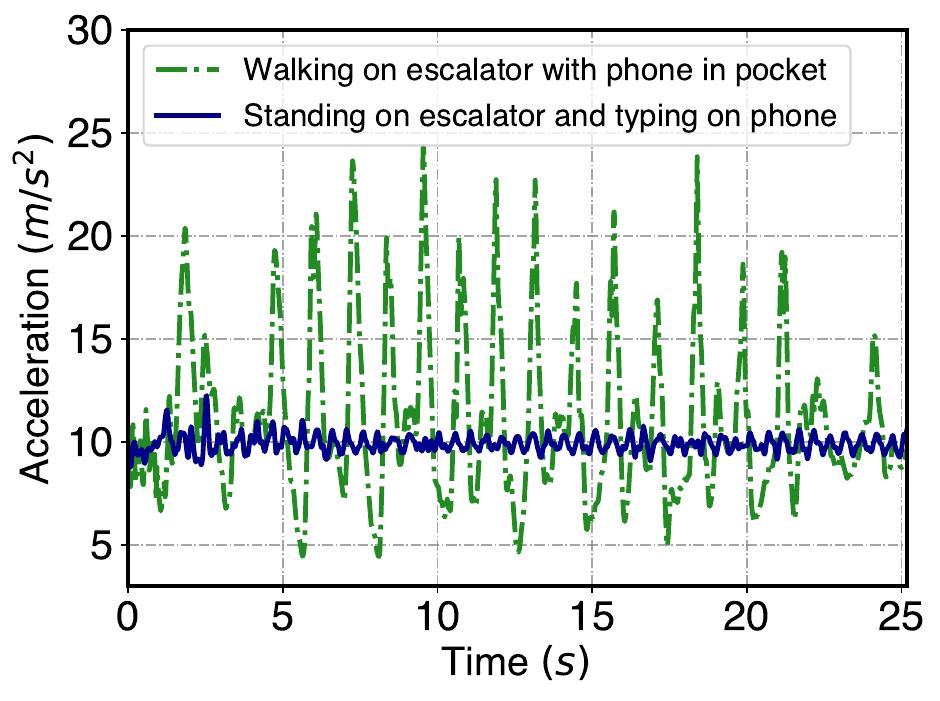}
        \caption{Illustration of the acceleration signals (in magnitude) in the ``escalator'' state under different pedestrian behaviors.  }  
    \label{fig:behavior_on_motion}
\end{figure}

When a pedestrian uses an elevator or escalator, the motion signal on the pedestrian's phone is mainly affected by two {\em independent} processes: the conveyor transport and pedestrian behavior. The conveyor affects the motion signal due to the transportation process; meanwhile, the pedestrian's various behaviors, whether consciously or unconsciously, more directly influence the phone movement, rotation, pose variation, etc. As a result, the motion signal reflects the mixture of the two processes, and due to various random pedestrian behaviors, the fragile signal of the conveyor transport is difficult to recognize, as illustrated in Figure~\ref{fig:behavior_on_motion}. 

We formulate the generation process of a motion signal under conveyor state ($s$) and pedestrian behavior ($V_p$) as a function, denoted as $G_m$, defined by 
\begin{equation}
    G_m(s, V_p, V_u)=x_m,     
    \label{eq: motion_generation}
\end{equation}
where the variable $V_u$ represents the minor unobserved factors to ensure the rigor of the equation. As mentioned earlier, our goal is to extract a feature of conveyor state that is independent of the pedestrian behavior. However, we neither have the labels of pedestrian behavior ($V_p$) nor the expression of the inverse function of the generation process ($G^{-1}_m$).\footnote{In other words, handcrafting reliable features of conveyor state under arbitrary pedestrian behaviors could be extremely difficult. }

\begin{figure}[tp]
    \centering
        \includegraphics[width=0.48\textwidth]{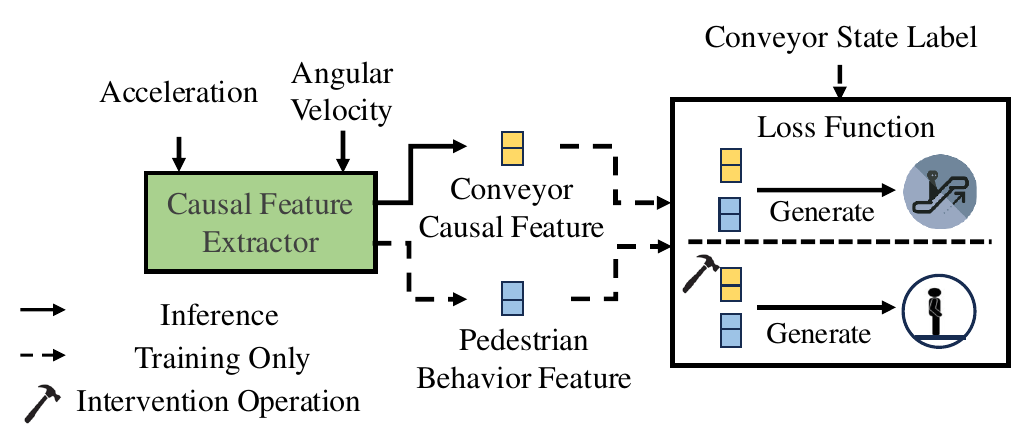}
        \caption{Causal feature extractor segregates conveyor state from pedestrian behavior based on causal decomposition, learned from a novel loss function. The key idea of the loss function is to simulate an intervention operation in deep feature space.}  
    \label{fig: motion_extractor}
\end{figure}

To overcome this, our key idea is to build a deep learning model to conduct a {\em causal decomposition} on motion signals. Specifically, the model decomposes a motion signal into two deep features that separately encode the conveyor state and pedestrian behavior, such that the feature of conveyor state is independent of pedestrian behavior. As illustrated in Figure~\ref{fig: motion_extractor}, we use a {\em causal feature extractor}, denoted as $f_{\theta_m}$, to decompose the motion signals ($x_m$) into a conveyor causal feature ($z_c$) and a pedestrian behavior feature, denoted as $z_p$, expressed as 
\begin{equation}
    f_{\theta_m}(x_m)=[z_c, z_p],  
    \label{eq: causal_extractor}
\end{equation}
where the feature extractor is parameterized by $\theta_m$. In the model training, the causal feature extractor learns from a loss function designed for extracting the {\em causal feature} of elevator and escalator.  
In the implementation, the structure of the causal feature extractor is empirically determined, consisting of a two-layer ConvLSTM and two fully connected layers with ReLU as the activation function. We assume that other temporal models should also be suitable~\cite{liang2024foundation}. 

In the following, we discuss the causal feature of elevator and escalator in Section~\ref{subsubsec: causal}, and present the loss function in Section~\ref{subsubsec: loss function}. 

\subsubsection{Causal feature of elevator and escalator}
\label{subsubsec: causal}

The causal feature of an object reflects the physical property of the object. 
To name a classic example, 
temperature generally drops as altitude goes up.
Since the law of physics universally holds in most cases, a causal feature is typically more reliable and independent of confounding factors to reflect an object than a common statistical feature~\cite{scholkopf2021toward}. 

In our problem, we identify that the {\em signal pattern caused by transportation} is the causal feature of elevator and escalator. 
To elucidate, as long as an elevator or escalator transports a pedestrian (and his/her smartphone), it inevitably generates the unique transport pattern on the motion signal. Conversely, the transport pattern would not exist without such a transporting process. 
Therefore, the transport pattern is the causal feature of elevator and escalator. Furthermore, the pattern is naturally {\em independent} of pedestrian behaviors, because it only depends on the conveyor. 

Fundamentally, the transport pattern needs to be extracted from {\em interventional experiment}. The experiment contrasts pairs of motion signals, between which the conveyor state is treated as univariate, and other variables are strictly controlled. Formally, each {\em experimental signal}, generated by $G_m(s, V_p, V_u)$, is paired with a {\em control signal}, generated by $G_m(\text{do}(s), {V_p}', {V_u}')$. To ensure the conveyor state ($s$) as the univariate, we conduct an intervention operation on $s$, denoted as $\text{do}(s)$, which manipulates the conveyor state to be the ``neither'' state, and strictly control the other variables such that ${V_p}'=V_p$ and ${V_u}'=V_u$.\footnote{The application of the {\em do-calculus} follows~\cite{scholkopf2021toward,pearl2009causal}. }
Given the above, the transport pattern, or the causal feature, can be reflected by the difference between the two signals, shown as 
\begin{equation}
    \Delta x_m=G_m(s, V_p, V_u)-G_m(\text{do}(s), {V_p}', {V_u}').          
    \label{eq: causal_obj_1}
\end{equation}

The interventional experiment underlies the causal decomposition upon the experimental signal, which is shown as
\begin{equation}
    G_m(s, V_p, V_u) = \Delta x_m + G_m(\text{do}(s), {V_p}', {V_u}').      
    \label{eq: causal_obj_2}
\end{equation}
This decomposition is causal because it captures the causal feature of elevator and escalator.  
In particular, if the conveyor state ($s$) is the ``elevator'' or ``escalator'' state, $\Delta x_m$ reflects the transport pattern due to the conveyor; if not, $\Delta x_m$ would be a zero vector, indicating the absence of a conveyor. 

With sufficient signal pairs collected from the interventional experiments (or simply {\em interventional data}), we can learn the causal feature extractor based on Equation~(\ref{eq: causal_obj_2}). 
However, the interventional data, by and large, are inconvenient to collect due to the demanding univariate setting. 
In most cases, we only have the experimental signal (or {\em observational data}) without the control signal. 
Therefore, we present a loss function for learning the causal decomposition based on observational data.  

\subsubsection{Loss Function}
\label{subsubsec: loss function}
In the below, we simplify the conveyor state as binary in notation, referring to $s=1$ as either the ``elevator'' or ``escalator'' state and $s=0$ as the ``neither'' state.\footnote{This is only for simplifying the equation expressions. The possible values of the conveyor state are still the ``elevator'', ``escalator'' and ``neither''. } 

As mentioned earlier, the goal of the causal feature extractor, i.e. $f_{\theta_m}(x_m)=[z_c, z_p]$, is to causally encode the experimental signal, i.e. $G_m(s, V_p, V_u)$. Specifically, $z_c$ encodes the transport pattern, i.e., $\Delta x_m$, and $z_p$ encodes the control signal, i.e., $G_m(\text{do}(s), {V_p}', {V_u}')$. However, we do not have the control signal and hence the transport pattern as ground truth. To tackle this, we present three learning constraints to enforce the decomposition of the causal feature extractor to comply with Equation~(\ref{eq: causal_obj_2}), based on which we design a loss function. 

First, if the decomposition is causal, it should not cause information loss. In Equation~(\ref{eq: causal_obj_2}), the transport pattern and the control signal can be used to reconstruct the experimental signal. Accordingly, we design a loss function that enforces the decomposed features to reconstruct the experimental signal, shown as   
\begin{equation}
    \mathcal{L}_{rec}(\theta_m, \theta_g)=\sum_{x_m\in D} \text{MSE}\left( g_{\theta_g}(z_c+z_p+\sigma), x_m \right).    
    \label{eq: reconstruction_loss}
\end{equation} 
In the loss function, $\text{MSE}(\cdot, \cdot)$ calculates the mean squared error, $z_c+z_p$ is the vector addition between the two deep feature vectors,~$\sigma$ is a Gaussian noise empirically used to model the $V_u$ in Equation~(\ref{eq: motion_generation}), and $g_{\theta_g}(\cdot)$ is a signal generator parameterized by $\theta_g$. In the implementation, the signal generator has three fully connected layers which are trained jointly with the causal feature extractor.  

Second, if the decomposition is causal, an intervention operation should effectively remove the causal feature of the conveyor. In the interventional experiment, we carry out an intervention operation to remove the conveyor from the generation process of the experimental signal, which results in the control signal. To simulate this process in deep feature space, combining the conveyor causal feature and pedestrian behavior feature should be able to generate the experimental signal; also, when we remove the conveyor causal feature, the pedestrian behavior alone should generate the control signal. Rewriting this as a constraint gives 
\begin{equation}
\begin{aligned}
        g_{\theta_g}(z_c+z_p+\sigma)&=G_m(s, V_p, V_u),  \\
        g_{\theta_g}(z_p+\sigma)&=G_m(\mathrm{do}(s), {V_p}', {V_u}'), 
\end{aligned}
\label{eq: sec_loss_obj}
\end{equation}
recalling that $g_{\theta_g}(\cdot)$ is the signal generator. 
Although the control signal is unavailable, we know that it belongs to the ``neither'' state. 
This allows us to give a loose constraint of Equation~(\ref{eq: sec_loss_obj}): the pedestrian behavior feature alone should generate the signal whose distribution conforms to the ``neither'' state, which is shown as
\begin{equation}
    g_{\theta_g}(z_p)\sim \mathcal{D}(x_m\mid s=0),      
    \label{eq: intervention_obj_2}
\end{equation}
where $\mathcal{D}(x_m\mid s=0)$ represents the distribution of motion signal in ``neither'' state.  
Since the signal is generated from the feature, we directly implement this constraint in deep feature space, facilitated by a classifier denoted as $k_{\theta_k}(\cdot)$. The loss function is shown as 
\begin{equation}
\begin{aligned}
    \mathcal{L}_{sim}(\theta_m, \theta_k) = 
    \sum_{\left(x_n, \tilde{s}_n\right)  \in D} & \Bigg[ 
    \mathrm{CE}\left(k_{\theta_k}(z_c + z_p), s = \tilde{s}_n\right) \\ 
    & \quad + \mathrm{CE}\left(k_{\theta_k}(z_p), s = 0\right)  \Bigg],
\end{aligned}
\label{eq:intervention_loss}
\end{equation}
where $\mathrm{CE}(\cdot)$ is the cross-entropy loss function, and recall that $\tilde{s}_n$ is the conveyor state label. In the implementation, the classifier has two fully connected layers which are trained jointly with the causal feature extractor.  

Third, if the decomposition is causal, the control signal should have a larger variance than the transport pattern. In Equation~(\ref{eq: causal_obj_2}), the control signal and the transport pattern are the results of causal decomposition. Since the {\em diverse} pedestrian behavior only affects the control signal and not the transport pattern, the variance of the control signal should generally be larger than that of the transport pattern. Formally, if the decomposition is causal, we should have  
\begin{equation}
    \text{Var}\left[ g_{\theta_g}(z_c) \right] < \text{Var}\left[ g_{\theta_g}(z_p) \right],
\end{equation}
where $\text{Var}[\cdot]$ calculates the variance of a vector. This constraint could be useful in learning the causal decomposition, as it complements the loose constraint in Equation~(\ref{eq: intervention_obj_2}). 
Similar to the implementation in Equation~(\ref{eq:intervention_loss}), we directly transform this constraint into a loss function enforced in deep feature space. Specifically, we reduce the variance of the causal feature, and the loss function is presented as 
\begin{equation}
\mathcal{L}_{con}(\theta_m) = \sum_s \sum_{\substack{(x_n, \tilde{s}_n) \in D, \\ \tilde{s}_n = s}} \text{Var}(z_c \mid \tilde{s}_n). 
\label{eq: consist_loss}
\end{equation}

In summary, the loss function of the causal feature extractor is given as 
\begin{equation}
    \mathcal{L}_{Cal}= \mathcal{L}_{sim} + w_1\mathcal{L}_{rec} + w_2\mathcal{L}_{con}. 
    \label{eq: cf loss}
\end{equation}
where $w_1$ and $w_2$ are the weights for tuning their relative importance. 
In the implementation, we learn the causal feature extractor using this loss function end-to-end. 

\begin{figure}[tp]
    \centering
        \includegraphics[width=0.3\textwidth]{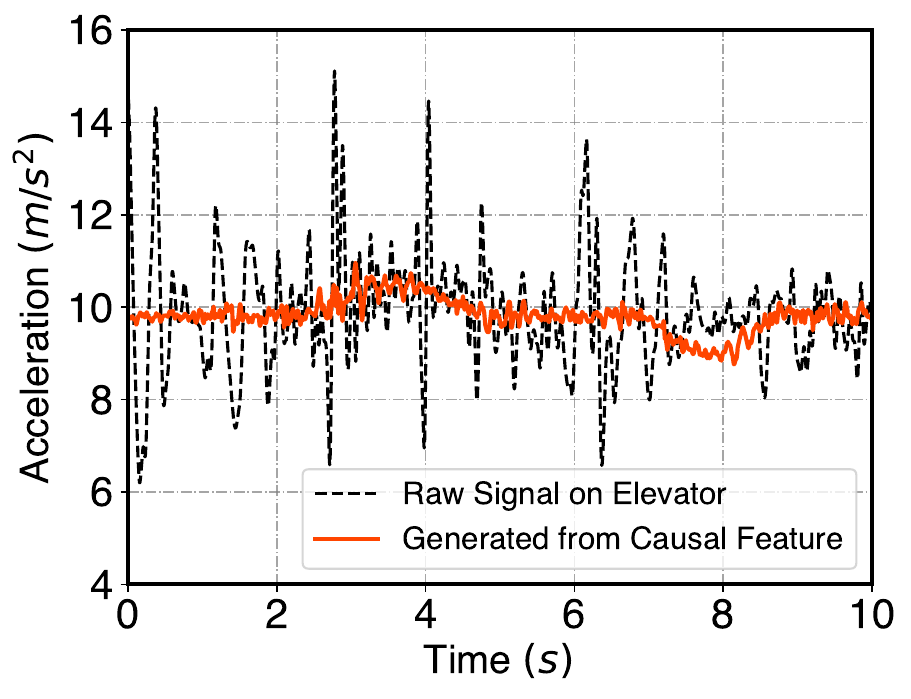}
        \caption{Illustration of acceleration signal under elevator ascending. }  
    \label{fig:interpret}
\end{figure}

Finally, we provide an analysis to interpret the causal feature extractor. Figure~\ref{fig:interpret} shows an example of the acceleration process of elevator ascending, where the process starts at the 2{\em nd} second and ends at the 9{\em th} second. The raw acceleration signal is very noisy because the pedestrian perturbs the signal in the process, performing behaviors such as typing, shaking, and moving. This makes the pattern of elevator transport very difficult to recognize. In comparison, the signal generated from the causal feature (aided by the signal generator in Equation~\ref{eq: reconstruction_loss}) demonstrates the pattern of elevator transport: it first enforces an ascending acceleration, followed by a descending one to maintain a stable speed at the end of the process. This provides an empirical understanding that causal feature enhances the classification of conveyor states. 

More rigorously, the causal feature extractor can be regarded as the {\em importance sampling} on training data to reduce the bias caused by pedestrian behaviors~\cite{tokdar2010importance}. Applying Bayes' theorem, the classification without the loss function can be written as 
\begin{equation}
\begin{aligned}
    P(s\mid x_m)&= \frac{P(x_m\mid s)P(s)}{P(x_m)}  \\
                &=\int P(s\mid V_p)\frac{P(x_m\mid s, V_p)}{P(x_m\mid V_p)}\quad dV_p,
\end{aligned}
\label{eq:bayes}
\end{equation}
where the {\em classification decision} on the left-hand side depends on the distribution of training data on the right-hand side. In other words, the decision depends on not only the motion signal but also the correlation between the conveyor state and pedestrian behavior, i.e., $P(s\mid V_p)$. As mentioned earlier, the conveyor states of pedestrians are largely independent of their behaviors in practice; however, such independence is {\em not} guaranteed when the training data are observational.\footnote{Note that such spurious correlation does not occur in interventional data. } As a result, the classification decisions are often {\em biased} due to the correlation in training data. For example, the classifier may misinterpret the browsing action as a feature of the ``escalator'' state when browsing frequently coincides with escalators in training data, which may cause errors when a pedestrian browses his/her phone outside an escalator. 

To address this, the proposed loss function enforces the conveyor state independent of pedestrian behavior in the training process. This can be interpreted as the importance sampling upon training data distribution. Specifically, the classification decision, after the importance sampling, would only depend on the motion signal, expressed as
\begin{equation}
\begin{aligned}
        P(s\mid x_m)&=\int wP(s\mid V_p)\frac{P(x_m\mid s, V_p)}{P(x_m\mid V_p)}\quad dV_p\\
        &=\int \frac{P(s\mid V_p)}{P(s\mid V_p)} \frac{P(x_m\mid s, V_p)}{P(x_m\mid V_p)}\quad dV_p\\
        & =\int \frac{P(x_m\mid s, V_p)}{P(x_m\mid V_p)}\quad dV_p, 
\end{aligned}
\end{equation}
where $w=1/P(s\mid V_p)$ represents the sampling weights for balancing the distribution. 

\subsection{Magnetic Feature Extractor}\label{subsec: magnetic}

\begin{figure}
    \centering
    \begin{minipage}[t]{.23\textwidth}
		\centering
		\includegraphics[width=\textwidth]{Figure/demo_mag_el.pdf}
    \end{minipage}
\hspace{0.01in}
 \begin{minipage}[t]{.23\textwidth}
		\centering
		\includegraphics[width=\textwidth]{Figure/demo_mag_es.pdf}
    \end{minipage}
        \caption{Examples of magnetic field intensity in elevator (left) and on escalator (right).  }
            \label{fig:mag_demo}
\end{figure}

\begin{figure}
    \centering
    \begin{minipage}[t]{.24\textwidth}
		\centering
		\includegraphics[width=\textwidth]{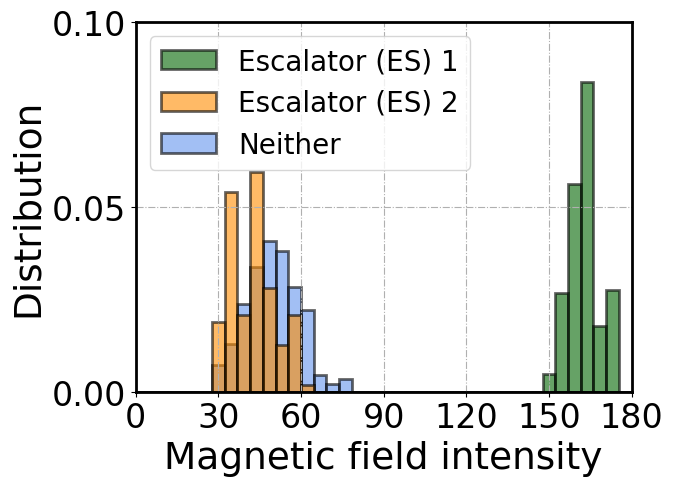}
    \end{minipage}
\hspace{0.01in}
 \begin{minipage}[t]{.23\textwidth}
		\centering
		\includegraphics[width=\textwidth]{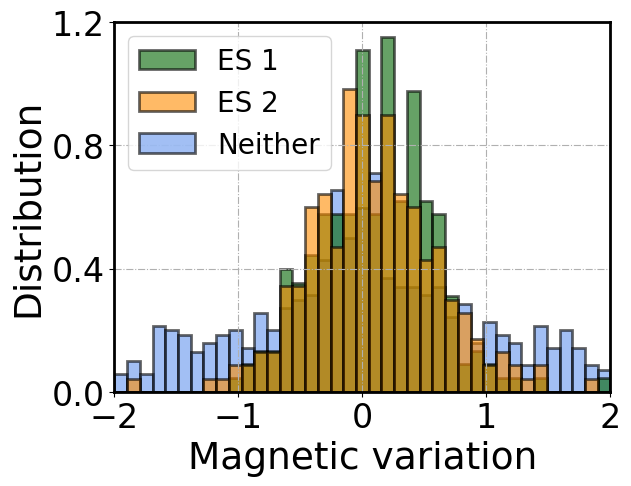}
    \end{minipage}
  \caption{Evidence that the temporal differential of the magnetic signal (right) exhibits greater stability as a characteristic of the conveyor state compared to raw magnetic field signals (left).  }
    \label{fig: mag_int_demo}
\end{figure}

A magnetometer regularly samples the magnetic field orientation at the phone location, generating a sequence of magnetic field signals, or simply a {\em magnetic signal}, presented as 
\begin{equation}
    x_b = \left[ x_b^{(0)}, x_b^{(1)}, ..., x_b^{(T-1)} \right], 
\end{equation}
where the magnetic signal has $T$ samples over time. When a pedestrian uses a conveyor, the signal would vary according to the conveyor movement. Yet another impact is that the signal is sensitive to the metallic structure of the conveyor, either the enclosed shell of the elevator or the semi-open frame of the escalator. 
Therefore, the magnetic signal may be used to classify the conveyor state, complementing the motion signal, as examplified in Figure~\ref{fig:mag_demo}.  

However, it is not straightforward to classify conveyor states using magnetic signals. Since the magnetic field depends on location, as shown in Figure~\ref{fig: mag_int_demo} (left), the magnetic signals are difficult to be used for the classification in different places. In addition, the signal is affected by many pedestrian behaviors such as waving and rotating phones, which makes the signal noisy in practice. 

\begin{figure}[tp]
    \centering
        \includegraphics[width=0.5\textwidth]{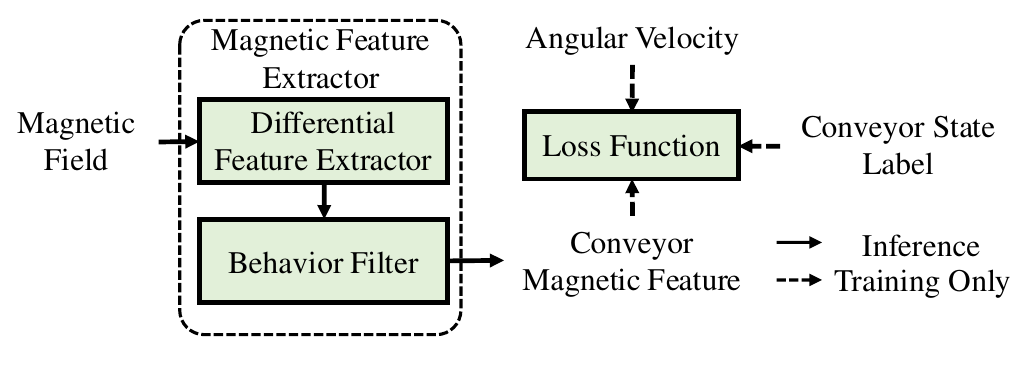}
        \caption{System diagram of magnetic feature extractor. 
        The differential feature extractor obtains from a magnetic signal a differential feature, based on which the behavior filter enhances the feature robustness against pedestrian behavior.  In training, the behavior filter learns from a loss function based on adversarial learning. }
    \label{fig: mag_extractor}
\end{figure}

To effectively leverage magnetic signals, we aim to extract a conveyor magnetic feature that is independent of conveyor locations and robust to pedestrian behaviors. 
As shown in Figure~\ref{fig: mag_extractor}, 
we propose a magnetic feature extractor consisting of a differential feature extractor and a behavior filter. 
Given a magnetic signal, the differential feature extractor extracts a {\em differential feature}, denoted as $\Delta x_b$, to alleviate the signal dependency on location. 
After that, a behavior filter reduces the signal noises due to pedestrian behaviors and outputs a conveyor magnetic feature. 

The design of the differential feature extractor is based on an empirical finding.  
As shown in Figure~\ref{fig: mag_int_demo}, compared with the raw magnetic signal, the temporal differential of the signal demonstrates much lower location dependency, and it maintains the ability to differentiate the conveyor states. This is because the differential feature of the signal reflects the magnetic variation of the conveyor process, which is much independent of their locations. 
Based on this observation, we design a {\em differential feature extractor}, denoted as $f_b(\cdot)$, which extracts the differential feature by
\begin{equation}
\begin{aligned}
 f_b(x_b) &= \Delta x_b \\
 & = \left[ \left| x_b^{(t)} \right|_2 - \left| x_b^{(t-1)} \right|_2, t = 1, 2, \ldots, T \right], 
\end{aligned}
\end{equation}
where $|\cdot|_2$ calculates the magnetic field intensity. Empirically, we find that the intensity is more effective than the orientation for reflecting conveyor states, which has been validated in Figure~\ref{fig: exp_mag}. 

To tackle the pedestrian behavior, we use a {\em behavior filter} to enhance the feature robustness against pedestrian behaviors. Formally, the behavior filter, denoted as $f_{\theta_b}(\cdot)$, takes a differential feature as input and outputs a conveyor magnetic feature ($z_b$), defined by 
\begin{equation}
    f_{\theta_b}(\Delta x_b)=z_b,  
    \label{eq: variation_feature}
\end{equation}
where the filter is parameterized by $\theta_b$. 
In the model training, we train the behavior filter robust against the perturbation caused by pedestrian behaviors. Specifically, we regard a behavior as a perturbation when it causes a phone to move or rotate, such as swinging and browsing, denoted as $V_p=1$, and otherwise, $V_p=0$. To enhance the robustness, we reduce the discrepancy of the conveyor magnetic feature between the two cases, which is shown as 
\begin{equation}
    \min \left| P(z_b\mid s=1, V_p=0)-P(z_b\mid s=1, V_p=1)\right|. 
    \label{eq: adv_goal}
\end{equation}
To enforce this, we employ a classifier, denoted as $k_{\theta_h}(\cdot)$, to determine the discrepancy between the two distributions. The classifier plays a min-max game with the behavior filter based on adversarial learning~\cite{lowd2005adversarial}, and the loss function for the behavior filter is shown as
\begin{equation} 
\mathcal{L}_{Mag}(\theta_b) = -\sum_{\substack{(x,\tilde{s}_n) \in D, \\ \tilde{s}_n = 1}} \mathrm{CE}\left(k_{\theta_h}\circ f_{\theta_b}(\Delta x_b), V_p \right), 
\label{eq: adv} 
\end{equation}
where $\circ$ is the composition operator. 

In the implementation, we use a threshold of angular velocity (1.5{\em rad/s}) to determine $V_p$ automatically. This is based on the observation that pedestrians could often lead to a high angular velocity, but elevators and escalators usually cannot. The behavior filter consists of a two-layer ConvLSTM and two fully connected layers with ReLU as the activation function.

\section{Evidential State Classifier}\label{sec: detector}

Given the extracted causal and magnetic features, 
\sysname{} employs an evidential state classifier to estimate the confidence of the conveyor states.  
We overview the classifier in Section~\ref{subsec: overview},  
and present its loss function in Section~\ref{subsec: detail}.

\subsection{Overview}\label{subsec: overview}
 
Given a conveyor state feature extracted in a period, say, 2 seconds, we aim to classify the conveyor state and estimate the confidence behind the classification decision. We trust the decision if the confidence is greater than a threshold and remain {\em undecided} (UD) otherwise.  
By discarding the decisions with low confidence as UD, we could enhance the system reliability. Therefore, it is important for the confidence to effectively reflect the classification accuracy.  

In conveyor state classification, the confidence depends on the INS signal due to the {\em signal-to-noise ratio} (SNR) and the {\em signal similarity} to the training data. 
For example, if pedestrian behavior is much more intense than the conveyor motion, resulting in low SNR, the INS signal could be ambiguous to reflect conveyor states, leading to low confidence. On the other hand, even with a high SNR, an INS signal may not be correctly classified if it is out of the training data of the classifier, which indicates low confidence. 
To account for the two aspects, existing frameworks are mainly based on the Bayes method and evidence theory~\cite{gawlikowski2023survey, josang2016subjective}.\footnote{The Softmax-based classifier is often overconfident as it does not consider the signal similarity~\cite{gawlikowski2023survey}. }

\begin{figure}[tp]
    \centering
        \includegraphics[width=0.41\textwidth]{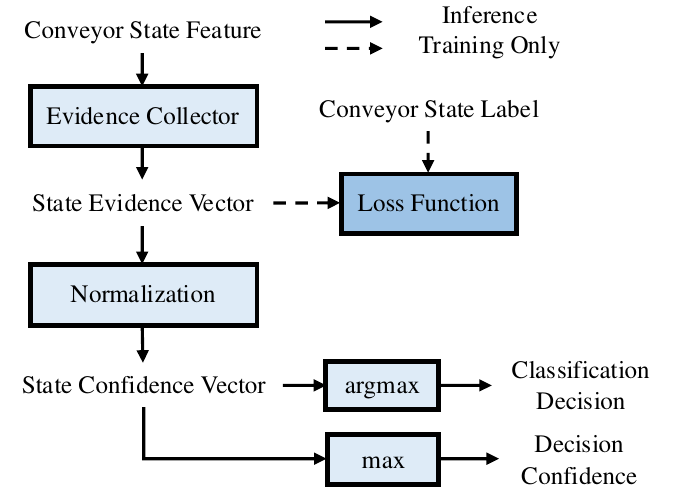}
        \caption{System diagram of evidential state classifier. Given a conveyor state feature, an evidence collector extracts the evidence supporting each of the states. The evidence is then normalized to be a state confidence vector that gives the classification decision and confidence.  
        }
    \label{fig: classifier} 
\end{figure} 

In this paper, we build an {\em evidential state classifier} due to computational efficiency.\footnote{Evidential classifier is as efficient as a Softmax-based classifier~\cite{sensoy_evidential_2018}. } The system diagram of the classifier is shown in Figure~\ref{fig: classifier}.
Given a conveyor state feature ($z$), it first uses an {\em evidence collector}, denoted as $f_{\theta_e}(\cdot)$, to extract a state evidence vector, denoted as $E$, defined by 
\begin{equation}
    f_{\theta_e}(z)=E, 
    \label{eq: collector}
\end{equation}
where $\{e_s\mid e_s\in E, e_s\geq 0\}$ is the evidence value supporting the conveyor state $s$ ($\text{dim}(E)=3$), and the evidence collector is parameterized by $\theta_e$. 
Then, it normalizes the evidence vector to be a {\em state confidence vector}, denoted as $C$, calculated as 
\begin{equation}
    C=\frac{E}{e_u+\sum_{e\in E}e}, 
    \label{eq: confidence}
\end{equation}
where $\{c_s\mid c_s\in C, c_s\in [0, 1)\}$ is the confidence of state $s$, $e_u$ is an uncertainty constant, and we empirically set it to $e_u=\operatorname{dim}(E)$. 
We select the state with the highest confidence to be the classification decision, which is given as
\begin{equation}
    s^*=\arg\max_{c\in C} c. 
\end{equation}
Finally, if the highest confidence is larger than a threshold, i.e. $\max(C)>\tau$, we trust the classification decision. Otherwise, the system outputs UD. 

In the implementation, the evidence collector consists of three fully connected layers. 
We use ReLU as both the activation function and the output layer, such that~$e_s\geq 0$. 
The evidence collector is supposed to extract evidence reflecting the two aspects of confidence.
In the following, we introduce the loss function for learning the evidence collector. 

\subsection{Loss Function}\label{subsec: detail}

Since the conveyor state of a pedestrian could only be one of the ``elevator,'' ``escalator,'' and ``neither'', the probabilities of the three states sum up to one. However, considering that the classifier may be unfamiliar with an INS signal (or the input signal is dissimilar to its training data), we use an {\em uncertainty term} to occupy a fraction of the probability. Specifically, the uncertainty term indicates the ``equally likely'' among the three states due to the unfamiliarity (namely the epistemic uncertainty~\cite{bao2021evidential}). By considering the uncertainty term, the probability assignment depends on the training data of the classifier. 
Therefore, the probability assigned to each of the conveyor states, i.e. $c_s\in C$, is called the {\em state confidence}. 
Formally, the uncertainty term, denoted as $u$, forms a simplex with the state confidence, which is shown as
\begin{equation}
    u+\sum_{c_s\in C} c_s=1. 
    \label{eq: class_goal}
\end{equation}

In evidence theory, confidence comes from {\em evidence}. The more evidence a classifier collects to support a decision, the less uncertainty it remains. 
Formally, recall that $e_s$ is the evidence value supporting the state $s$, and $e_u$ is the uncertainty constant, the uncertainty term can be calculated as
\begin{equation}
    u=\frac{e_u}{e_u+\sum_{e\in E}e}.     
\end{equation}
To estimate the evidence value, we learn an evidence collector, and the learning has two objectives. 

The first objective is to optimize {\em classification accuracy}. In other words, the evidence collector should maximize the evidence value supporting the state of ground truth while minimizing the values for the other states. Recalling that $x$ is the INS signal, $\mathrm{CE}(\cdot, \cdot)$ is the cross-entropy loss function, and $\tilde{s}_n$ is the conveyor state label, the loss function of the first objective can be written as
\begin{equation}
    \mathcal{L}_{cls}(\theta_m, \theta_b, \theta_e)=\sum_{(x, \tilde{s}_n)\in D} {CE}\left(\frac{E}{\sum_{e\in E}e}, \tilde{s}_n \right),
    \label{eq: ec}
\end{equation}
where the loss function applies to both the feature extraction module and the evidence collector, and this implicitly captures the signal SNR in confidence. 

The second objective is to optimize the {\em uncertainty term}. In other words, the evidence collector should extract more evidence values for its familiar signals, and vice versa. 
To make the uncertainty term differentiable, we transform the simplex in Equation~(\ref{eq: class_goal}) into a Dirichlet distribution whose variance positively relates to the uncertainty term.\footnote{Proof of equivalence is in~\cite{josang2016subjective}. }  
Let $a_s=e_s+1$ and $A=\sum_{e\in E}(e+1)$, the variance of the Dirichlet distribution after the transformation is 
\begin{equation}
    \text{Var}(E)=\sum_{s} \frac{a_s(A-a_s)}{A^2(A+1)}. 
    \label{eq:dirich_var}
\end{equation}
To optimize the uncertainty term, we minimize the variance of the distribution on training data. 
Overall, the loss function for learning the evidence collector is 
\begin{equation}
    \mathcal{L}_{Els}=\mathcal{L}_{cls}+\sum_{D} \text{Var}(E). 
\end{equation}

In summary, 
recalling that $\mathcal{L}_{Mag}$ and $\mathcal{L}_{Cal}$ are the loss functions of magnetic and causal feature extractors, respectively, 
the loss function of the whole system is given as
\begin{equation}
    \mathcal{L}=\mathcal{L}_{Els}+w_3\mathcal{L}_{Mag}+w_4\mathcal{L}_{Cal},  
    \label{eq: overall_loss}
\end{equation}
where $w_3$ and $w_4$ are their weights. In the implementation, we learn the classifier and the two feature extractors end-to-end.

\section{Illustrative Experimental Results}\label{sec: exp}

This section evaluates the performance of \sysname{}.  We first introduce the experimental setting in Section~\ref{subsec: exp_set}, and 
present the performance of \sysname{} in Section~\ref{subsec: exp_comp}.  Then, we study system parameters in Section~\ref{subsec: result}, 
and show efficiency studies in Section~\ref{subsec: exp_on_device}. 

\begin{figure*}[tp]
    \centering
    \begin{minipage}[t]{.34\textwidth}
		\centering
		\includegraphics[width=\textwidth]{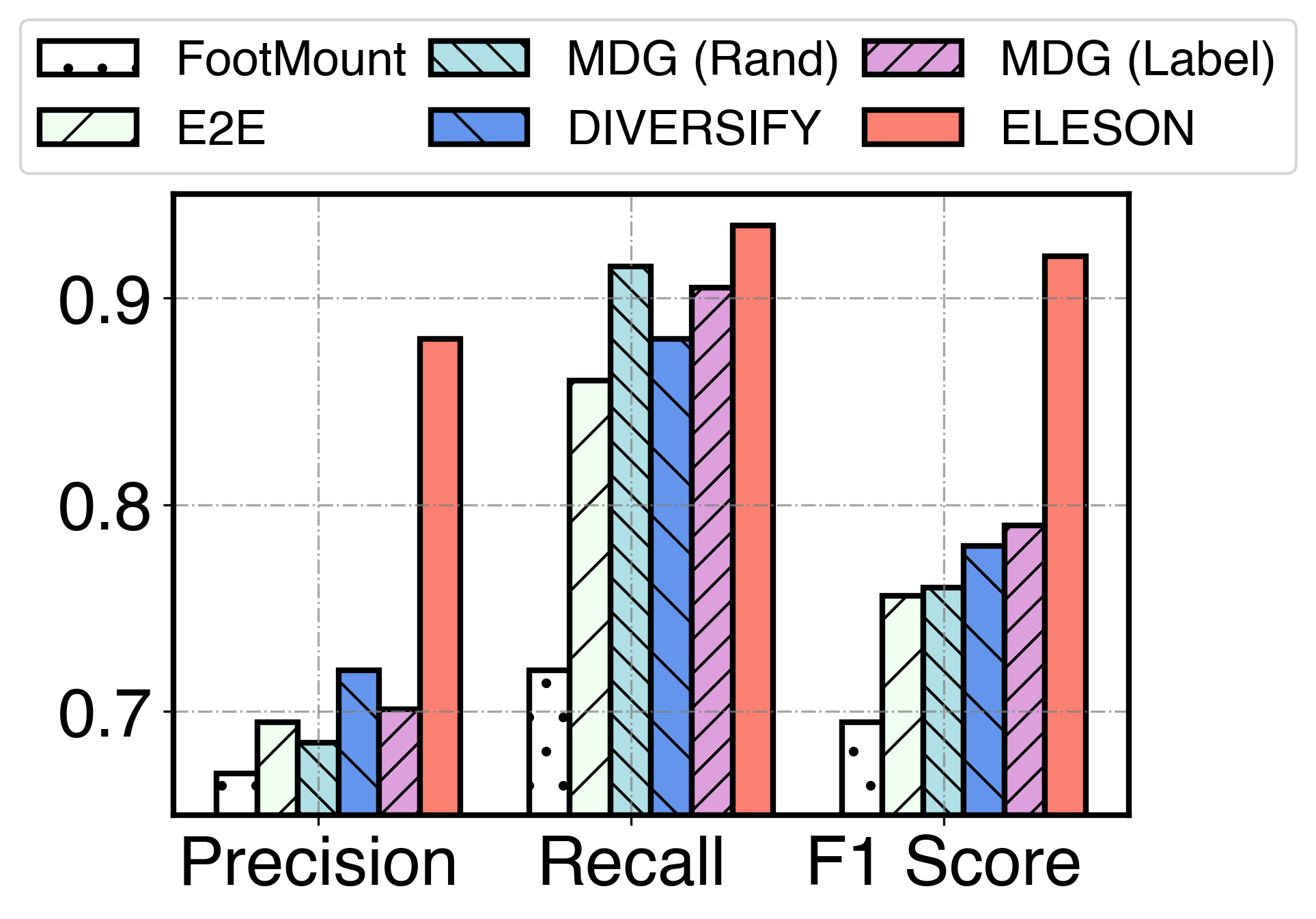}
		\caption{Under arbitrary pedestrian behaviors, \sysname{} outperforms previous approaches significantly, improving around 14\% in F1 score. }
		\label{fig: exp_accu}
	\end{minipage}
 \hspace{0.1in}
	\begin{minipage}[t]{.28\textwidth}
		\centering
		\includegraphics[width=\textwidth]{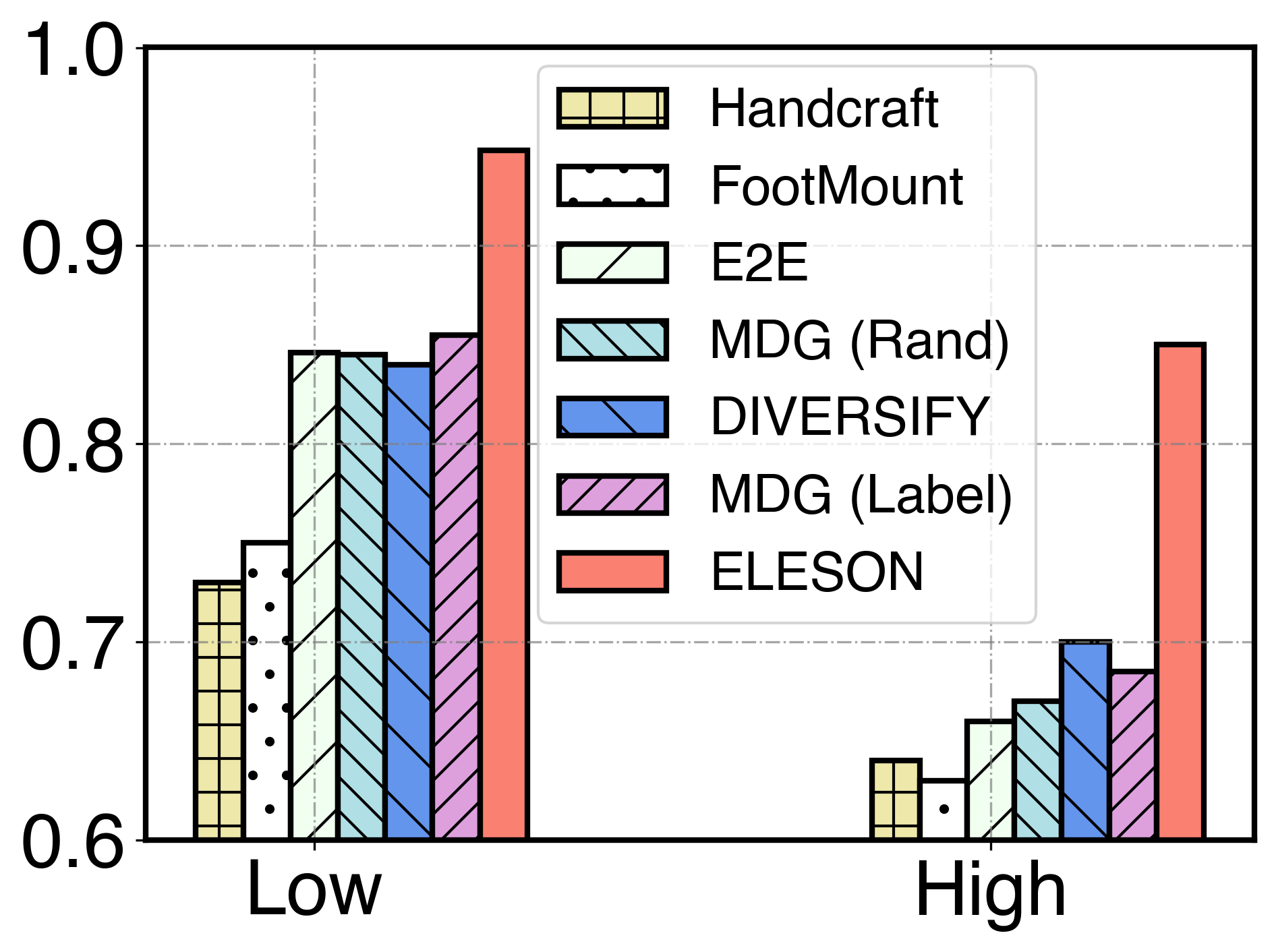}
		\caption{Under low and high levels of behavior perturbations, \sysname{} attains a satisfactory F1 score. }
		\label{fig: exp_gyro_accu}
	\end{minipage}
 \hspace{0.1in}
    \begin{minipage}[t]{.32\textwidth}
		\centering
		\includegraphics[width=\textwidth]{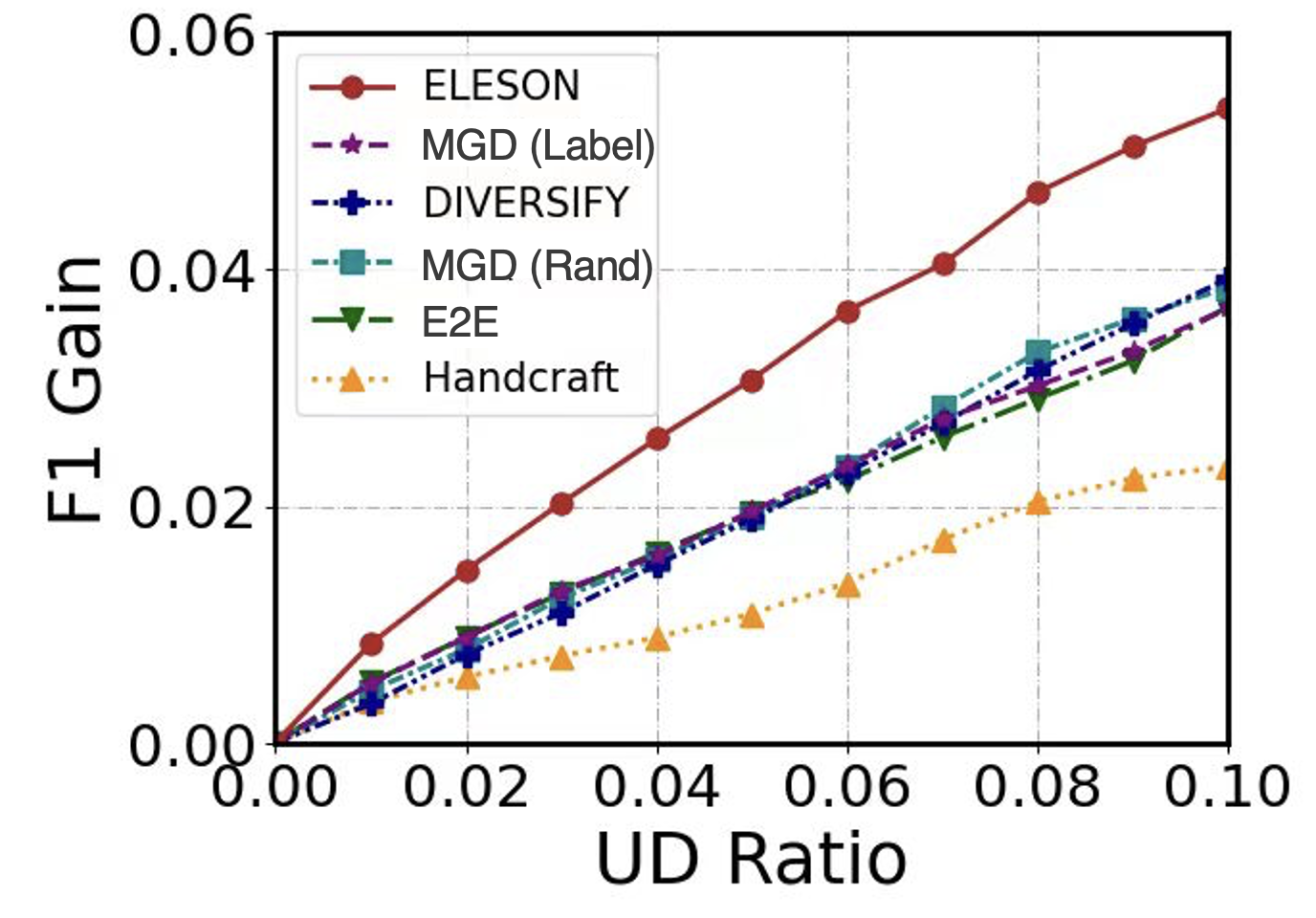}
		\caption{\sysname{} benefits higher F1 score from the same UD ratio due to the effective confidence estimation. }
		\label{fig: exp_f1_gain}
	\end{minipage}
\end{figure*}

\subsection{Experimental Setting}
\label{subsec: exp_set}

We curate a large and diverse collection of real-world data to validate \sysname{}. In the data collection, pedestrians freely roam shopping malls with arbitrary behaviors and casually carry their phones, during which they take elevators and escalators at different locations.\footnote{Note that the data are collected from the personal phones of the participants, covering different brands and models. } 
At the same time, an observer annotate the conveyor state with timestamps whenever the pedestrians get on and off the conveyors (with their consent). In total, we collect data from multiple pedestrians in 10 shopping malls over 20 hours, with roughly 20\% in elevators, 20\% on escalators, and 60\% are neither. To our knowledge, this is the first dataset for classifying the conveyor states of pedestrians under arbitrary behaviors based on phone INS.  

Since pedestrians could use conveyors for varied periods, we classify the conveyor state using a short sliding window with size and stride of 2 seconds, leading to 36,420 instances for classification in total. To balance conveyor state labels, we follow object detection and evaluate the performance using F1 score separately for ``elevator'' and ``escalator'' states, shown as
\begin{equation}
        \operatorname{F1\ score}=\frac{2\times \operatorname{Precision}\times \operatorname{Recall}}{\operatorname{Precision}+\operatorname{Recall}}.      
\end{equation}
Specifically, $\operatorname{Precision}=\operatorname{TP}/\left(\operatorname{TP}+\operatorname{FP}\right)$, and $\operatorname{Recall}=\operatorname{TP}/\left(\operatorname{TP}+\operatorname{FN}\right)$, where $\operatorname{TP}$, $\operatorname{FP}$, $\operatorname{FN}$ stand for ``true positive'', ``false positive'', and ``false negative'', respectively.    
Unless stated otherwise, we report the average F1 score over ``elevator'' and ``escalator'' states. 

\begin{figure*}[tp]
    \centering
       \begin{minipage}{0.48\textwidth}
        \raggedright
        \captionof{table}{F1 scores under various phone carriage (with the confidence threshold of 0). }
        \begin{tabular}{c|cccc} 
        \hline
            Carriage style  & Reading & In-pocket & Swing & In-bag \\
        \hline
            Handcraft & 0.73 & 0.66 & 0.67 & 0.60 \\
        \hline
            FootMount & 0.75 & 0.67 & 0.62 & 0.66 \\
        \hline
            E2E & 0.76 & 0.76 & 0.68 & 0.77 \\
        \hline 
            MDG~(Rand) & 0.79  &  0.75 & 0.71 & 0.73 \\
        \hline
            DIVERSIFY & 0.79 & 0.75 & 0.72  & 0.78 \\  
        \hline
            MDG~(Label) & 0.81 & 0.78 & 0.73 & 0.78 \\
         \hline
         \sysname{} &  0.92 &  0.92 &  0.84 &  0.88 \\
         \hline
        \end{tabular}
        \label{tb: style}
    \end{minipage}
\hspace{0.05\textwidth}
\begin{minipage}{0.41\textwidth}
        \centering
        \includegraphics[width=0.6\linewidth]{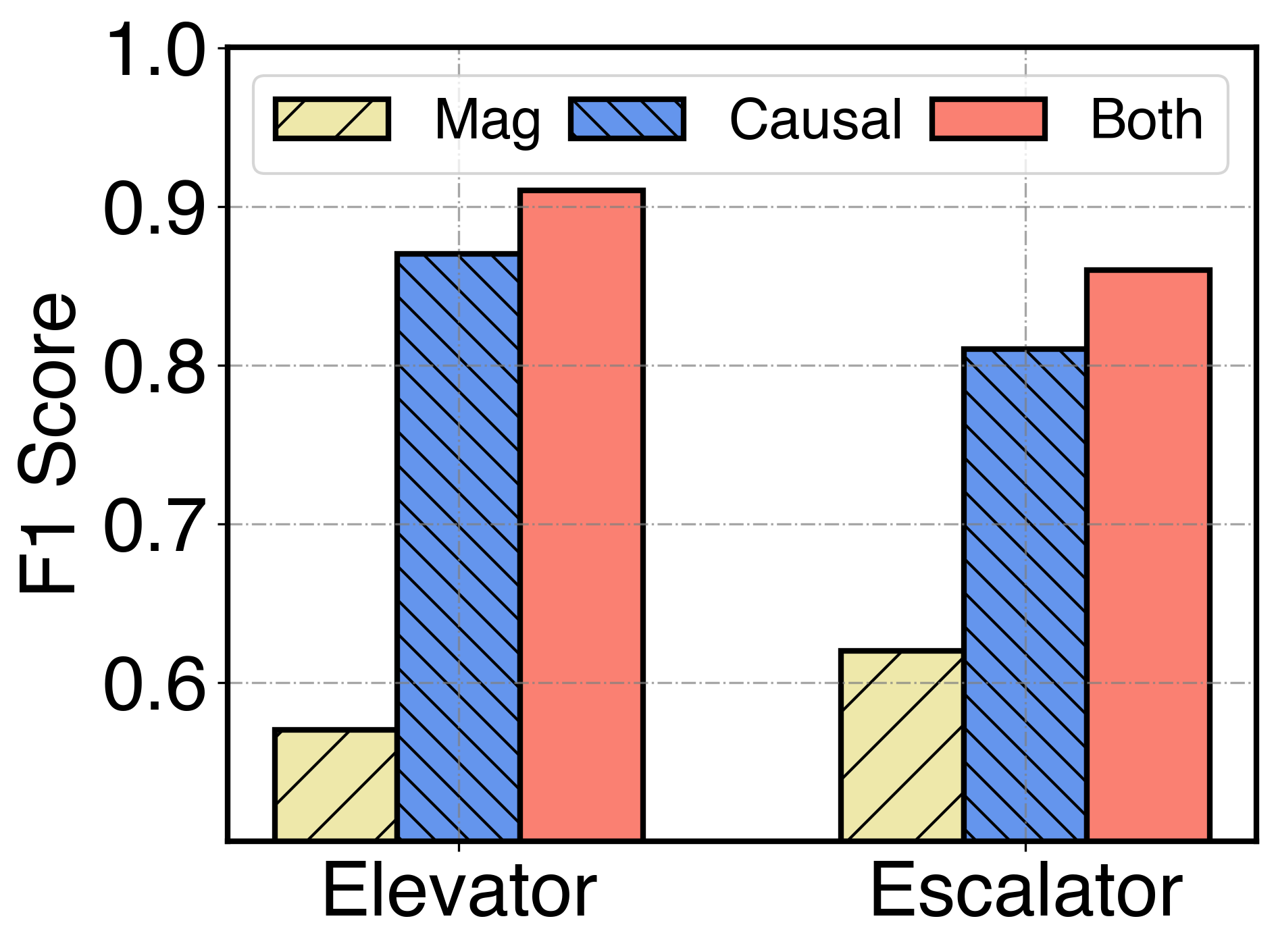} 
        \caption{Ablation study on conveyor causal and magnetic features (with the confidence threshold of 0). }
        \label{fig: exp_component}
    \end{minipage}%
\end{figure*}

To validate \sysname{}, we compare it with the following state-of-the-art approaches for conveyor state classification:  
\begin{itemize}[leftmargin=*]
    \item \emph{Handcrafted feature approach (Handcraft)}~\cite{liu_smartcare_2015} classifies conveyor states using several handcrafted features extracted from phone-based INS signals, assuming a steady holding posture. In the experiment, we use a neural network to classify the extracted features. 
    \item \emph{Foot-mounted sensor approach (FootMount)}~\cite{kronenwett2018elevator} classifies conveyor states using a finite-state machine specially designed for foot-mounted INS.  
\end{itemize}
To demonstrate the challenge of our problem, we further compare the performance with the following general classification approaches that have been extended to INS processing: 
\begin{itemize}[leftmargin=*] 
    \item \emph{End-to-end Approach (E2E)}~\cite{andrade2022human} 
        is our backbone approach, which classifies INS signal using ConvLSTM end-to-end.  
    \item \emph{Multi-domain Generalization (MDG)}~\cite{chen2023domain}
        enhances classification robustness using the labels of pedestrian behaviors.      
        To implement this, we label the behaviors using phone carriage styles (Label) and random grouping (Rand), shown in Table~\ref{tb: style}. The implementation is based on ConvLSTM. 
    \item \emph{DIVERSIFY}~\cite{lu_out--distribution_2023} enhances classification robustness based on implicit labeling, where it employs a clustering algorithm and reduces the feature variance among the clusters. In the implementation, we use ConvLSTM to classify conveyor states incorporating the scheme. 
\end{itemize}

We assess confidence estimation using the area under the receiver operating characteristic curve (AUROC), 
which measures the discriminability of confidence to distinguish correct and false classification decisions. 
We regard it as a true positive case when a false decision is regarded as UD. 
Specifically, AUROC measures the trade-off between false positive rate and true positive rate, which is computed as 
\begin{equation}
    \operatorname{AUROC}=\int_0^1 S_D(r) dr, 
\end{equation} 
where $r$ is the false positive rate and $S_{D}(r)$ maps $r$ to the true positive rate given a dataset $D$. 
We implement the following approaches to compare confidence estimation:
\begin{itemize}[leftmargin=*] 
    \item \emph{Entropy approach (Softmax)}~\cite{krizhevsky2012imagenet}
        uses information entropy of the classification score to present the confidence of Softmax-based approach; 
    \item \emph{Bayesian neural network (Bayesian)}~\cite{thiagarajan2021explanation} 
        represents confidence using prediction variance over the distribution of network parameters. 
        In the experiment, we sample network parameters using the Dropout mechanism~\cite{krizhevsky2012imagenet};
    \item \emph{Temperature scaling (TempScale)}~\cite{karandikar2021soft} 
        calibrates the classification scores using an exponential parameter (calibrated in training) 
        and calculates confidence using the information entropy of the scores. 
\end{itemize}

\begin{figure*}[tp]
    \centering
    \begin{minipage}[t]{.23\textwidth}
		\centering
		\includegraphics[width=\textwidth]{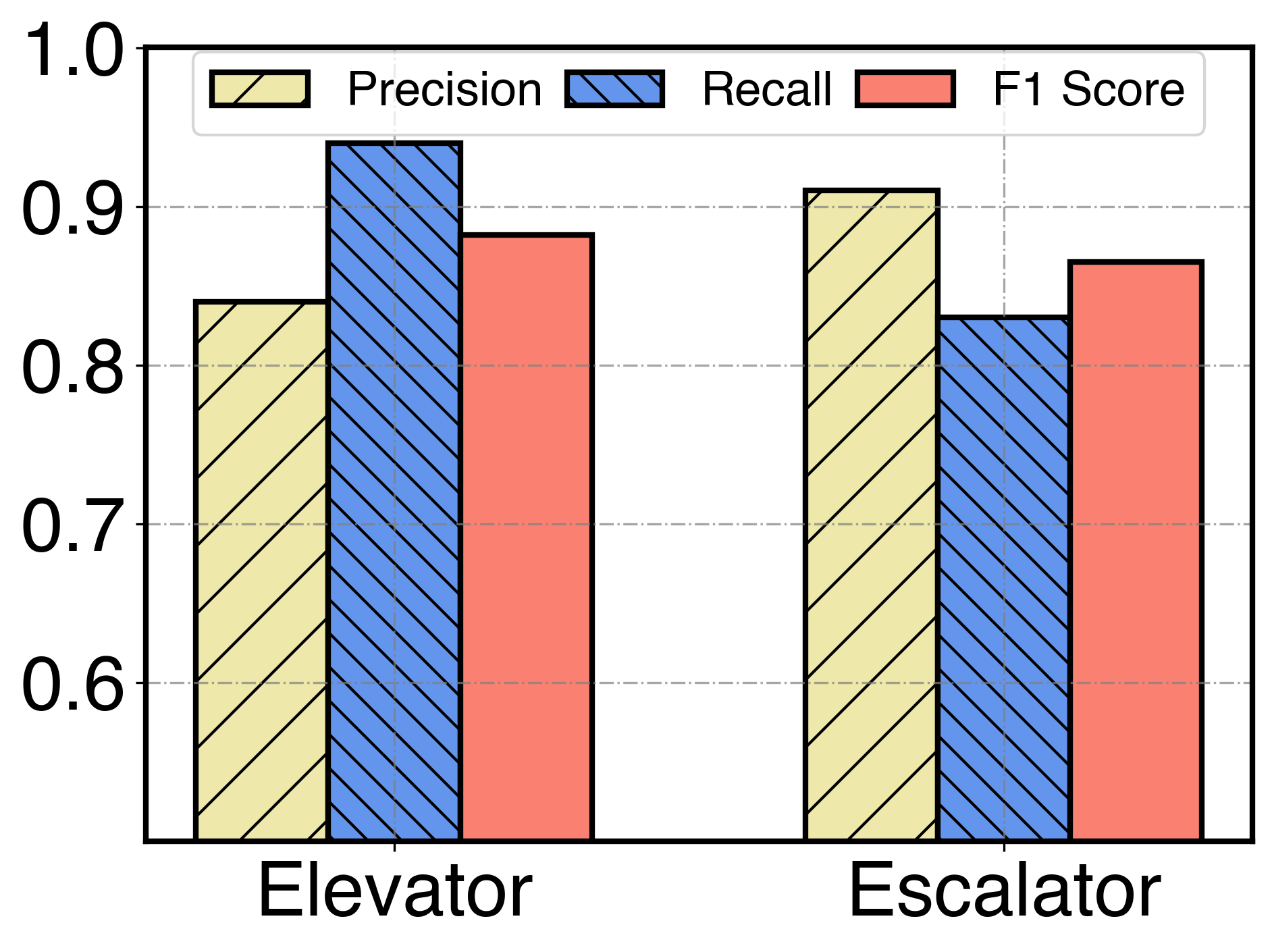}
		\caption{Performance on unseen elevators and escalators. }
		\label{fig: exp_crowd}
	\end{minipage}
    \hspace{0.1in}
 \begin{minipage}[t]{.23\textwidth}
		\centering
		\includegraphics[width=\textwidth]{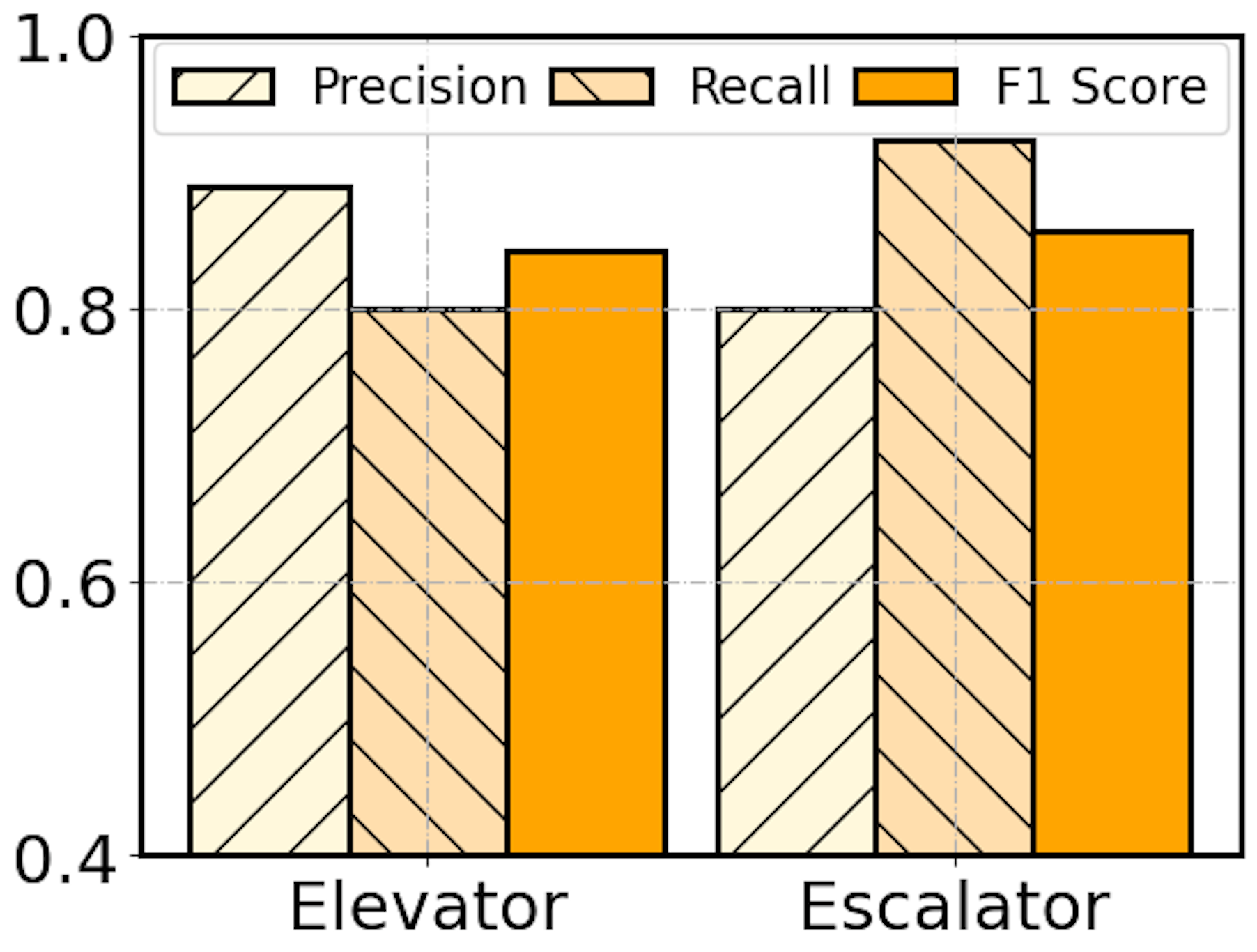}
		\caption{Performance on crowdsourced data. } 
		\label{fig: exp_crowdsource}
	\end{minipage}
  \hspace{0.1in}
 \begin{minipage}[t]{.23\textwidth}
		\centering
		\includegraphics[width=\textwidth]{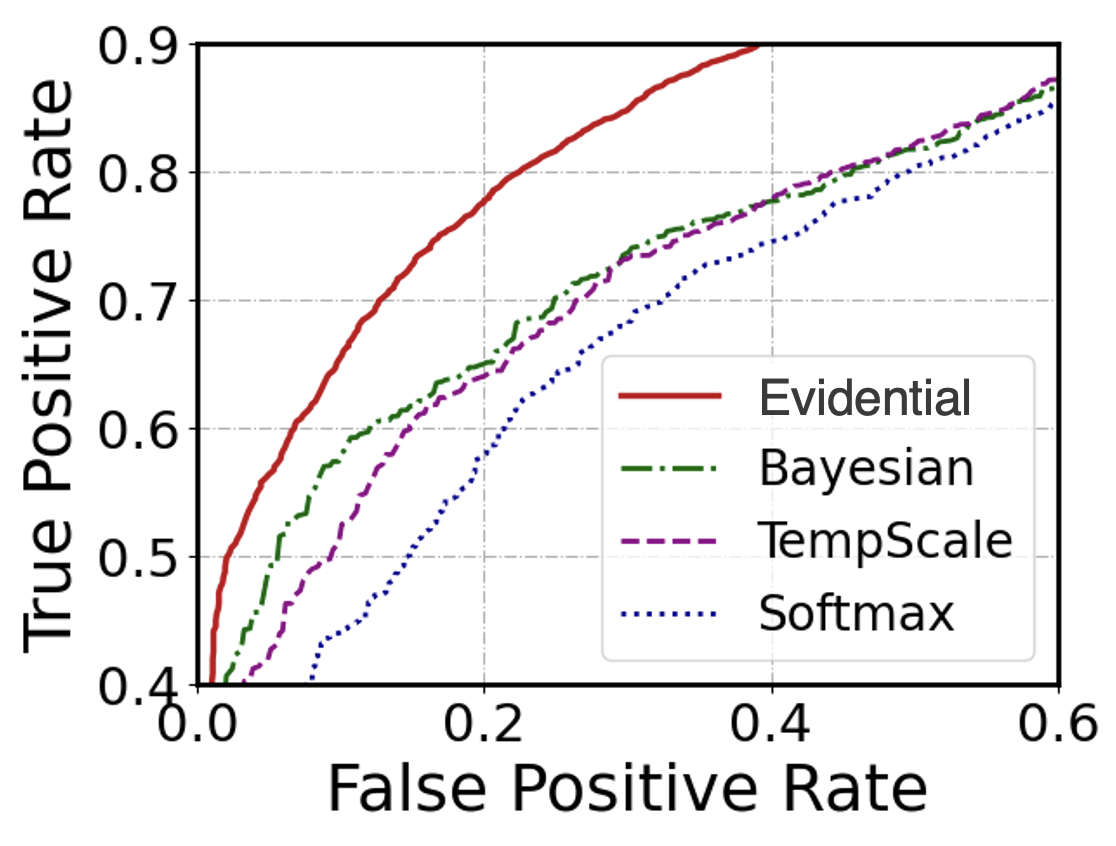}
		\caption{Study on confidence estimation in AUROC. } 
		\label{fig: exp_roc}
	\end{minipage}
 \hspace{0.1in}
	\begin{minipage}[t]{.23\textwidth}
		\centering
		\includegraphics[width=\textwidth]{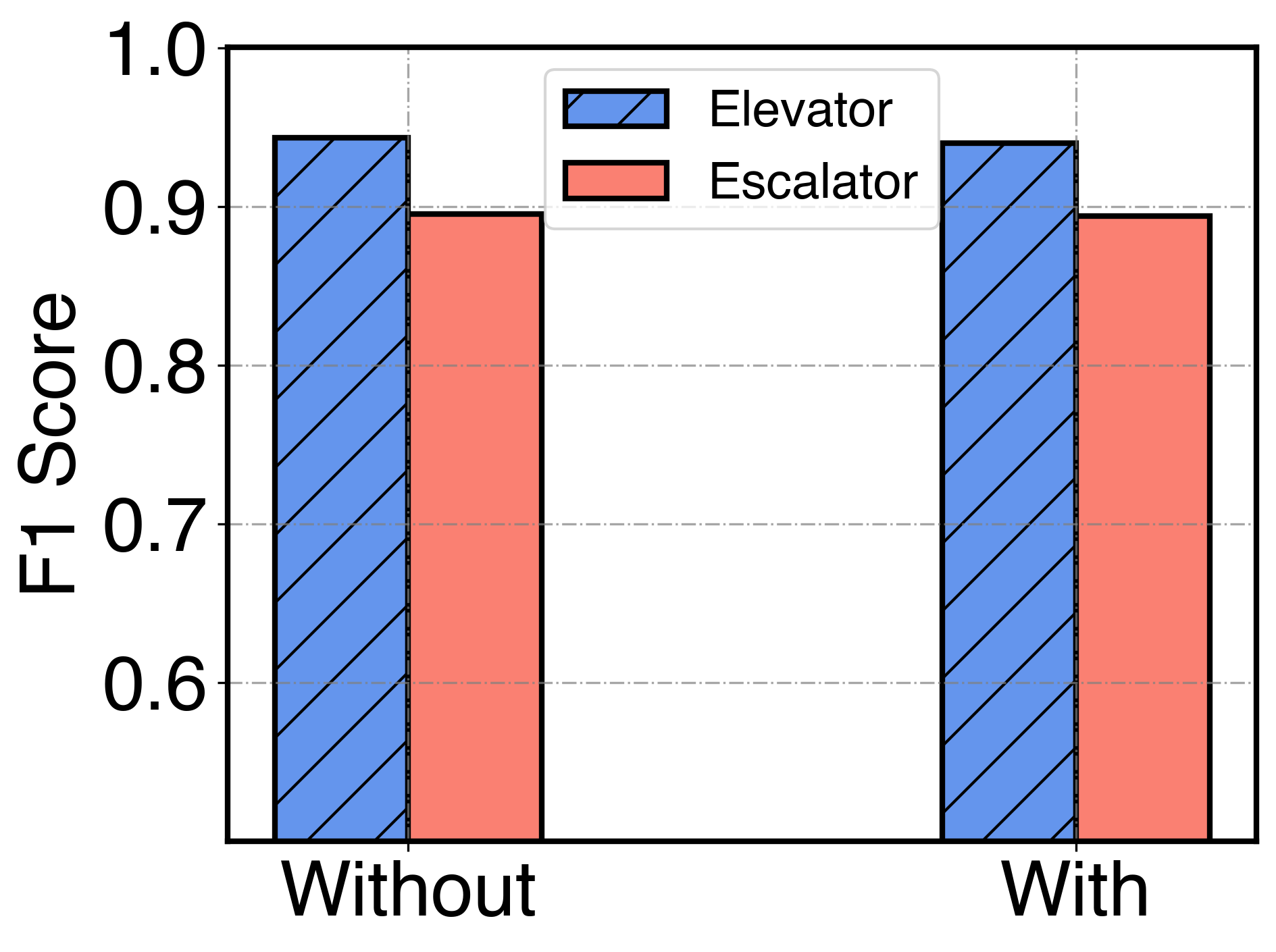}
		\caption{Ablation study on the uncertainty optimization. }
		\label{fig: exp_add_dirich}
	\end{minipage}
\end{figure*}

Furthermore, we have evaluated the computing efficiency of \sysname{} by deploying it to a mobile phone (Huawei LDN-AL10), based on which we investigated its memory usage, inference time, and power consumption (see Section~\ref{subsec: exp_on_device}).  

In the experiment, each of the conveyor causal feature and the magnetic feature has 128 dimensions, and the model is optimized by the Adam optimizer. The INS sampling frequency is 100{\em hz}, which is supported by most smartphones~\cite{androiddev,iosdev}. We empirically follow the signal preprocessing in~\cite{herath2020ronin}, accounting for the general heterogeneity issues, and we assume that other approaches should also be applicable~\cite{stisen2015smart,ru2022mems}. 
To reduce randomness, we use the five-fold cross-validation to evaluate the results, where we shuffle the signal sequences of each data collection.

\subsection{Overall Performance}
\label{subsec: exp_comp}

We compare the overall performance of the schemes on the whole dataset in Figure~\ref{fig: exp_accu}. From the figure, the previous classification approaches for conveyor states fail in our setting due to their behavior or sensor assumptions (note that {\em Handcraft} performs similarly to {\em FootMount}). On the other hand, the general classification approaches cannot achieve satisfactory results on our problem, which validates that our problem is new, open, and challenging. 
In comparison, \sysname{} has achieved satisfactory precision, recall, and hence F1 score (around 0.89) owing to the feature extraction module.  
In addition, by setting the confidence threshold to be 0.5, which leads to the UD ratio of 0.05, \sysname{} gains around 3\% improvement in F1 score, while others are less than 2\%. With this setting, \sysname{} has achieved 0.92 in F1 score, improving at least 14\% compared to the previous approaches. 

Figure~\ref{fig: exp_gyro_accu} shows the performance of \sysname{} under different levels of behavior perturbation, where the level is classified by a threshold of angular velocity (1.5 rad/s), and around 40\% instances are of the high level. Compared with their performance in the low level of perturbation, all schemes, including \sysname{}, degrade under the high level. This is because the behavior perturbation undermines the signal-to-noise ratio. Specifically, the general classification approaches, as they leverage neural networks, perform better than the previous approaches for conveyor state classification; however, their performance is not satisfactory, because they lack sufficiently precise labels of pedestrian behaviors. In comparison, \sysname{} shows superior F1 scores over 0.85 in both levels without the need for any behavior labels. 

Figure~\ref{fig: exp_f1_gain} shows how the F1 score varies with the UD ratio, which is tuned by the confidence threshold. In the experiment, the comparison schemes use the Softmax-based classifier as in their original setting. Compared with the other schemes, \sysname{} gains significant improvements in F1 score under the same UD ratio. This validates that the confidence estimation of \sysname{} can further enhance the system reliability. Therefore, we set the confidence threshold as 0.5, which leads to 3\% improvement in F1 score with merely the UD ratio of 0.05. 

Phone carriage style is a coarse but explainable way to label pedestrian behaviors.
Table~\ref{tb: style} shows the F1 scores of the comparison schemes under different phone carriage styles with the confidence threshold of 0 (or UD ratio of 0).  
From the table, all schemes perform relatively better in stable carriage styles (such as reading) than dynamic styles (such as swinging). However, the F1 score of the previous works is less than satisfactory under all phone carriage styles. This validates that they cannot achieve satisfactory results without precise labels of behaviors. In comparison, \sysname{} achieves a satisfactory F1 score ($\ge 0.84$) under various carriage styles, which is consistent with the previous results. 

Figure~\ref{fig: exp_component} shows an ablation study on the conveyor causal and magnetic features with the confidence threshold of 0. 
From the figure, the causal feature contributes to the majority of accuracy, and the magnetic feature further boosts the classification effectiveness. This is because they characterize the conveyor state of pedestrians from different aspects. 
Furthermore, the figure shows that the F1 score of the ``elevator'' state is higher than that of the ``escalator'' state. This is because, in our observation, pedestrians in elevators usually perform fewer actions than on escalators, thus introducing less perturbations. 

\begin{figure*}[tp]
    \centering
    \begin{minipage}[t]{.23\textwidth}
		\centering
		\includegraphics[width=\textwidth]{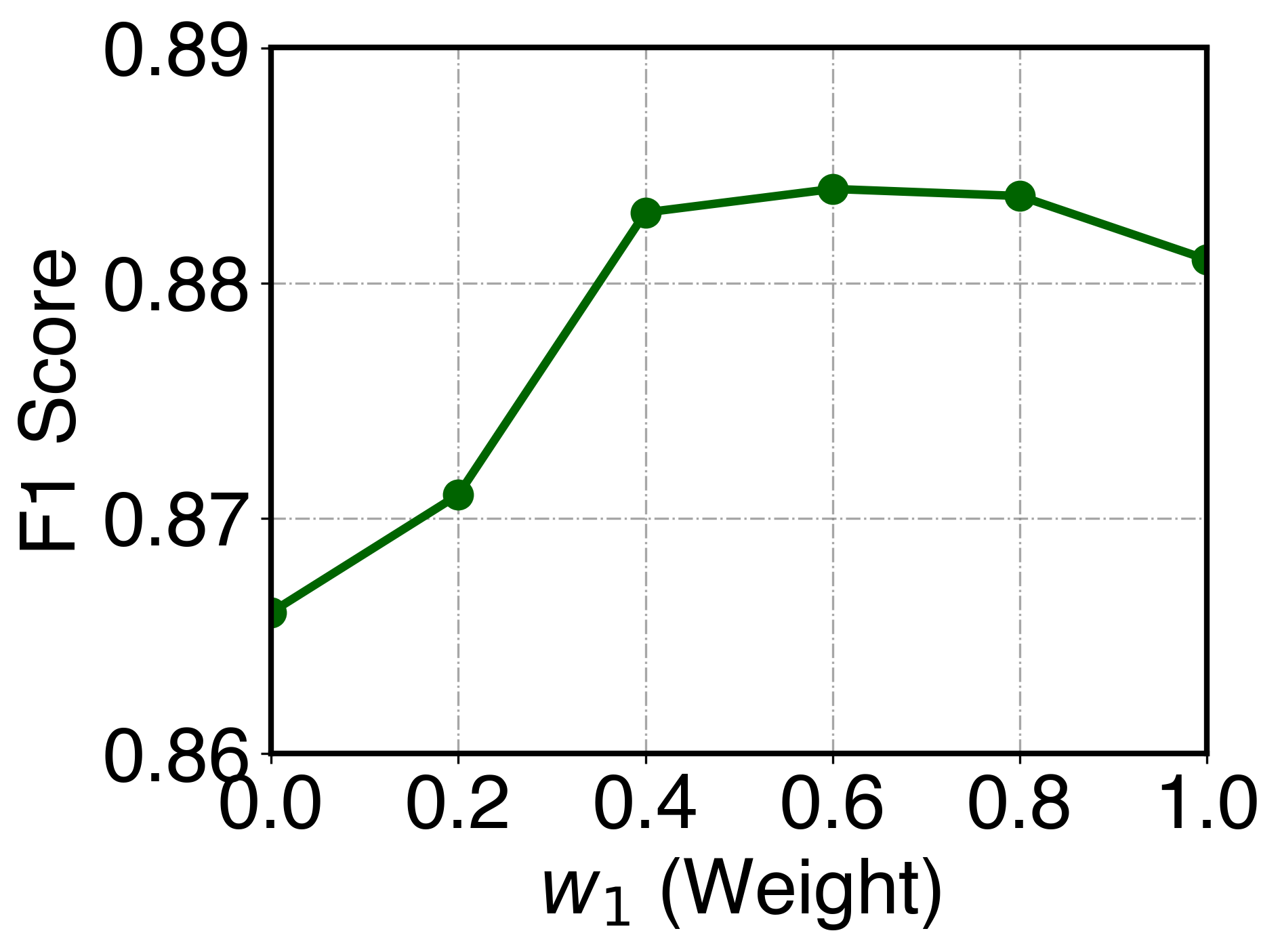}
		\caption{Parameter study on $w_1$ in the loss function for causal feature extractor. }
		\label{fig: exp_inv}
	\end{minipage}
    \hspace{0.1in}
    \begin{minipage}[t]{.23\textwidth}
		\centering
		\includegraphics[width=\textwidth]{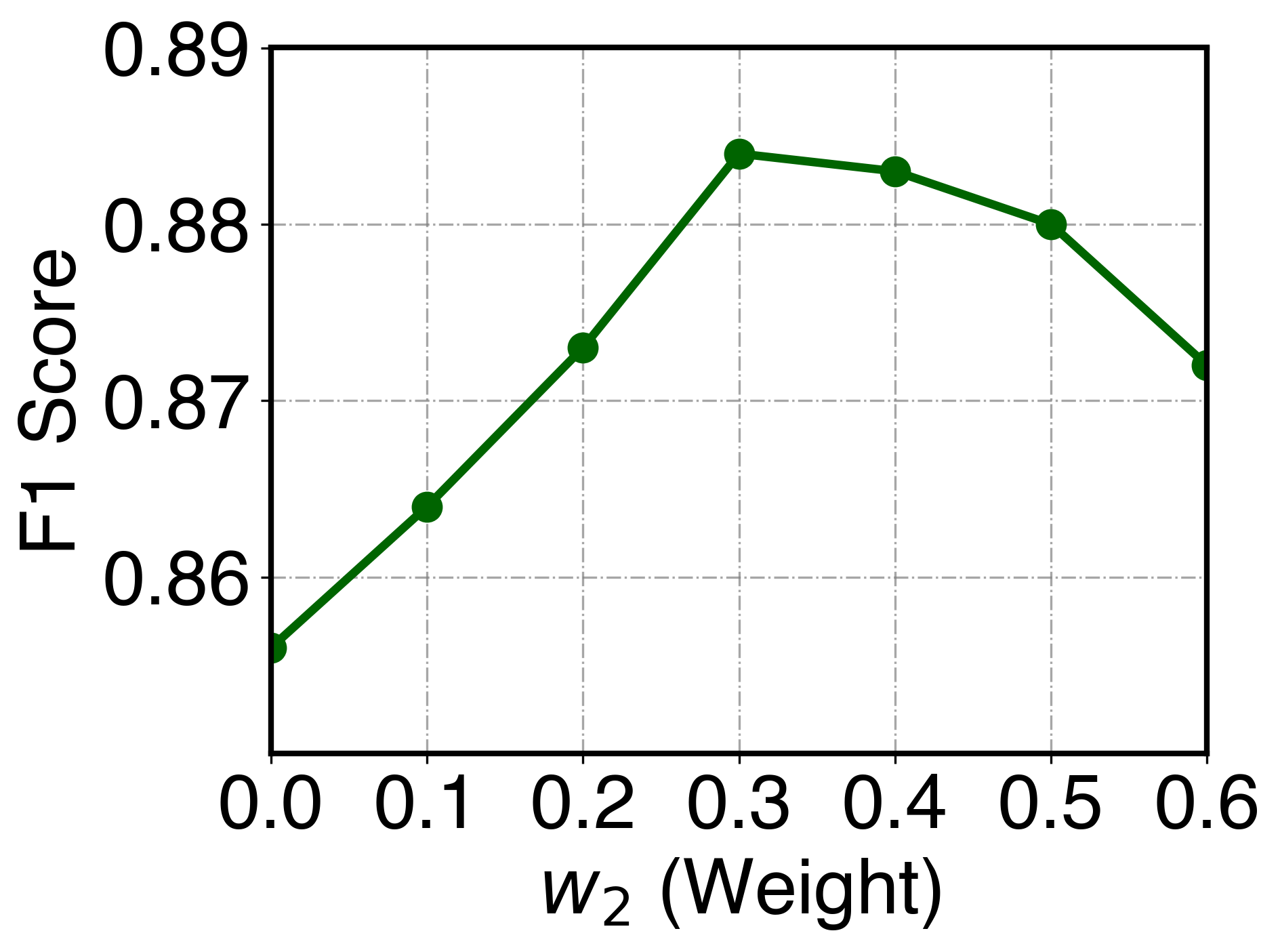}
		\caption{Parameter study on $w_2$ in the loss function for causal feature extractor. }
		\label{fig: exp_consist}
	\end{minipage}
     \hspace{0.1in}
  \begin{minipage}[t]{.23\textwidth}
		\centering
		\includegraphics[width=\textwidth]{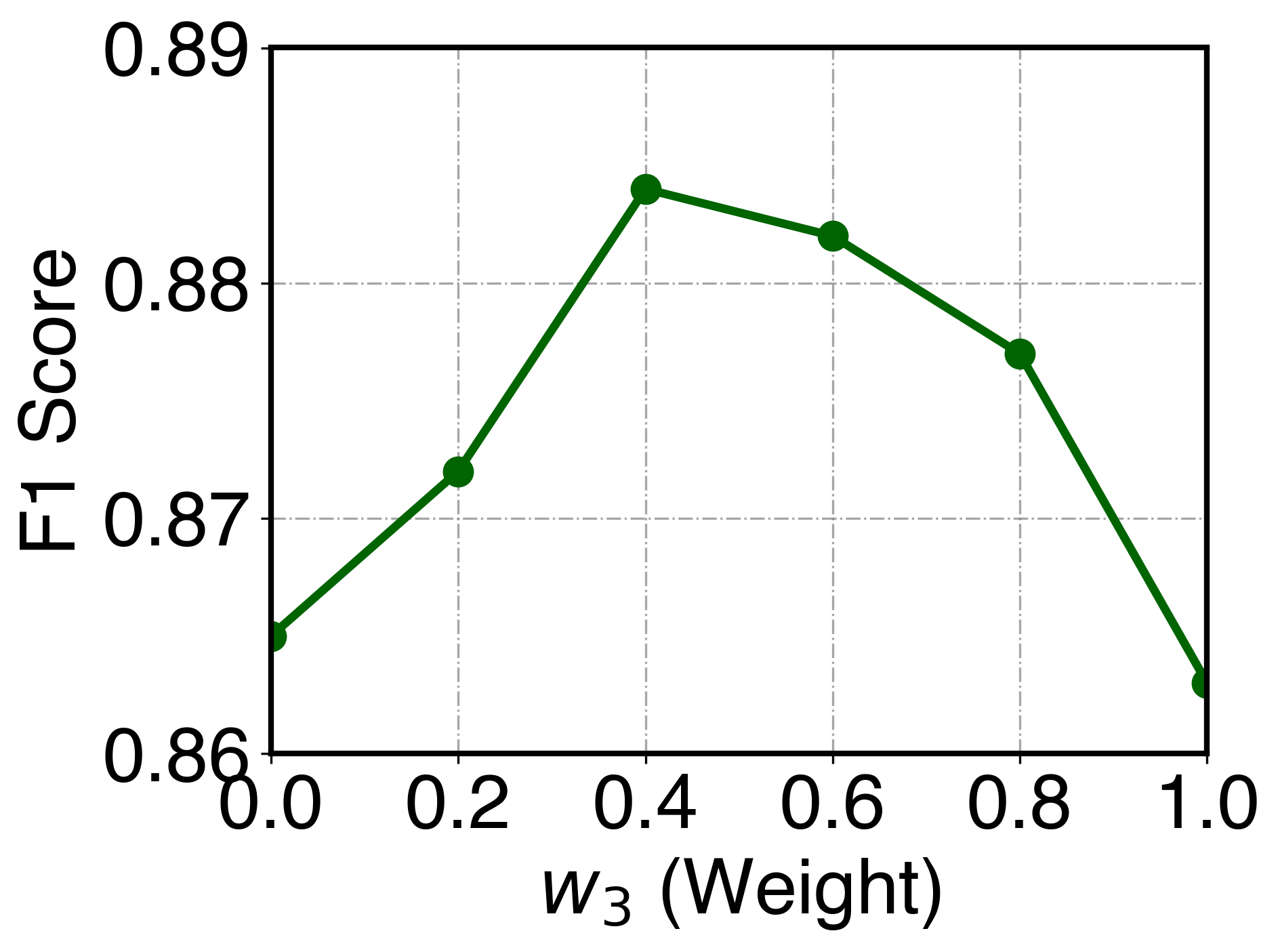}
		\caption{Parameter study on $w_3$ for the loss function of magnetic feature extractor. }
		\label{fig: exp_beh}
	\end{minipage}
    \hspace{0.1in}
   \begin{minipage}[t]{.21\textwidth}
		\centering
		\includegraphics[width=\textwidth]{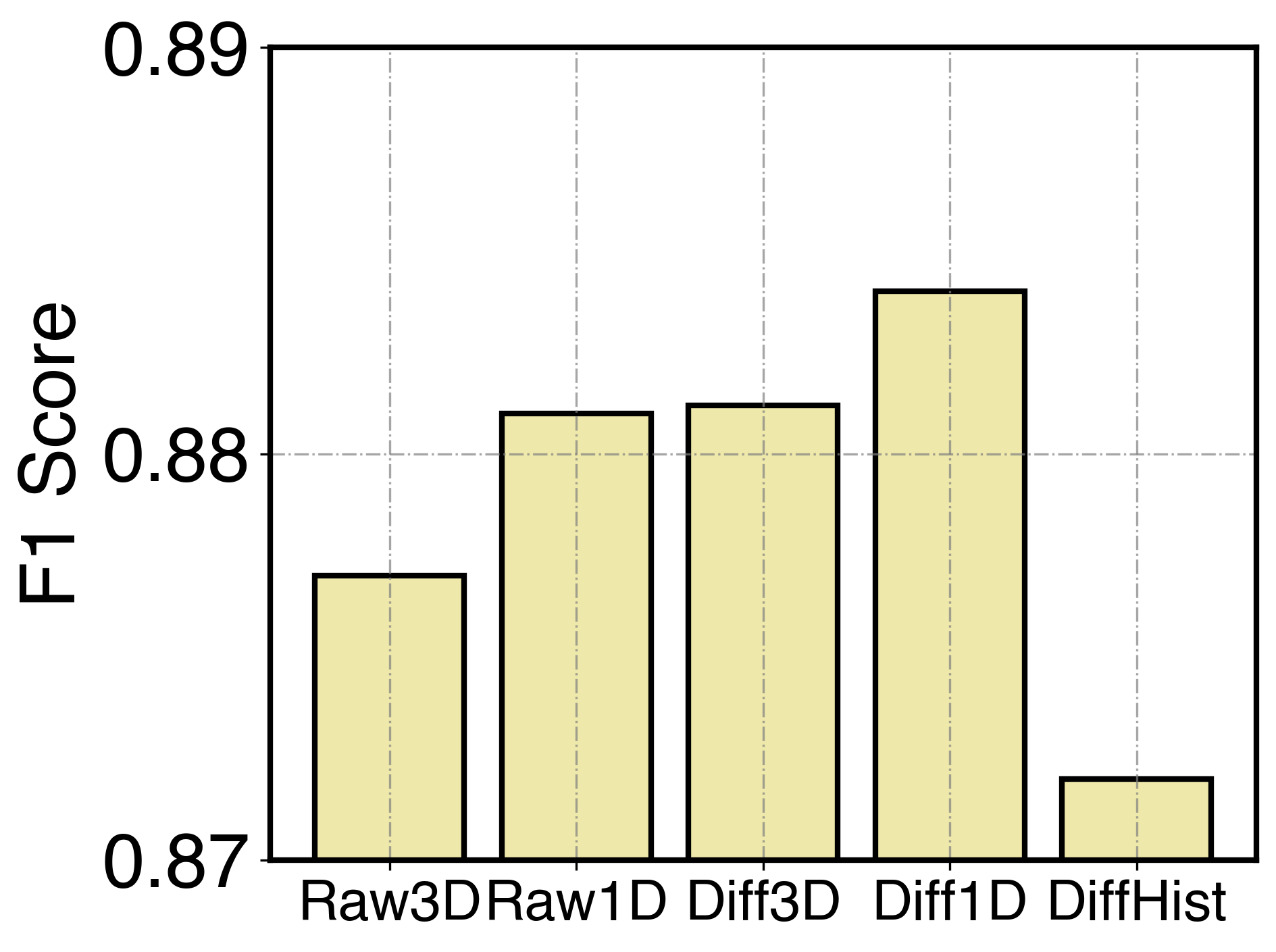}
		\caption{Ablation study on differential feature extractor.  }
		\label{fig: exp_mag}
	\end{minipage}   
\end{figure*}
\begin{figure*}[tp]
    \centering
    \begin{minipage}[t]{.22\textwidth}
		\centering
		\includegraphics[width=\textwidth]{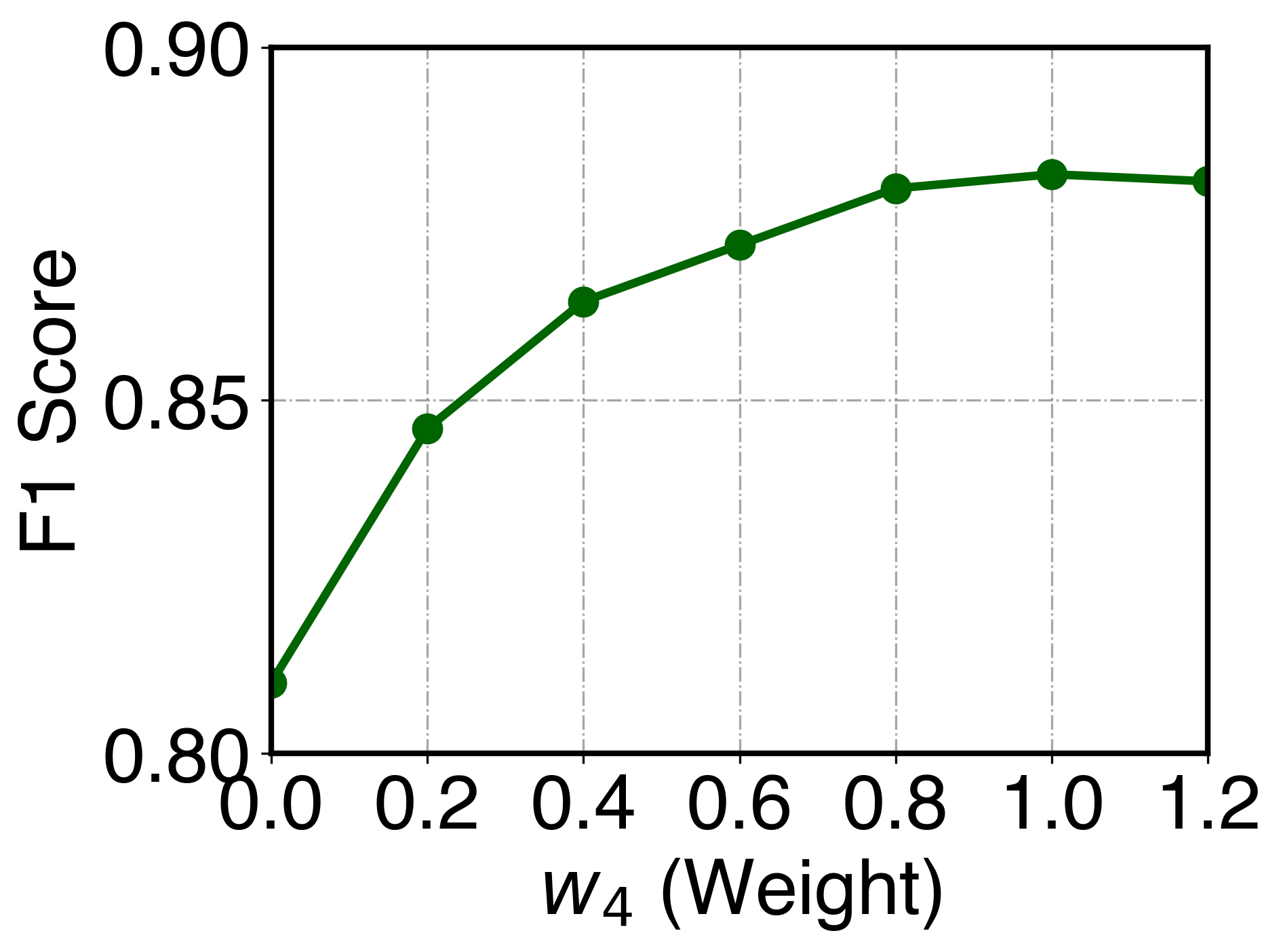}
		\caption{Parameter study on $w_4$ for the loss function of causal feature extraction.  }
		\label{fig: exp_w4}
	\end{minipage}   
    \hspace{0.1in}
	\begin{minipage}[t]{.21\textwidth}
		\centering
		\includegraphics[width=\textwidth]{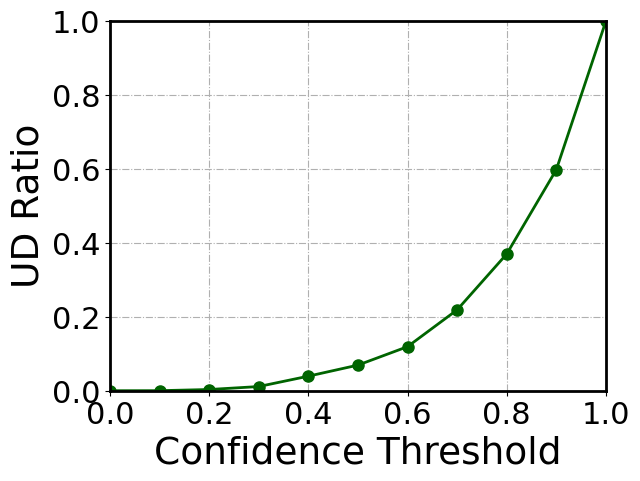}
		\caption{Distribution of confidence over the whole dataset. }
		\label{fig: exp_ud_conf} 
	\end{minipage}
    \hspace{0.1in}
    \begin{minipage}[t]{.24\textwidth}
		\centering
		\includegraphics[width=\textwidth]{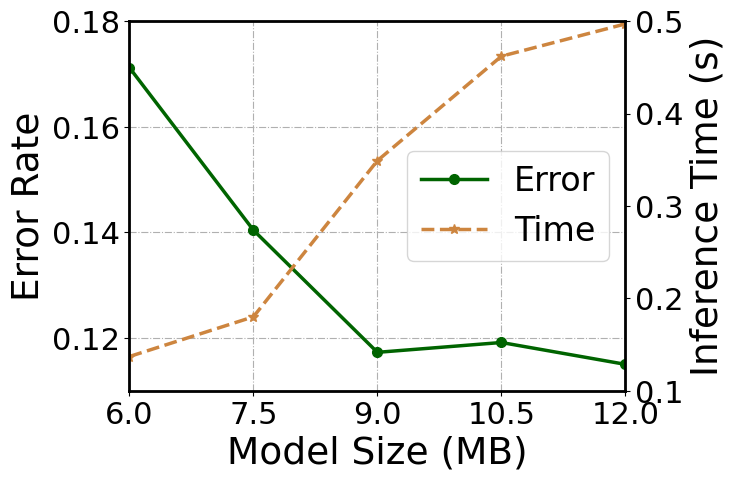}
		\caption{Inference time and error rate versus model size. }
		\label{fig: exp_edge}
	\end{minipage}
    \hspace{0.1in}
	\begin{minipage}[t]{.21\textwidth}
		\centering
		\includegraphics[width=\textwidth]{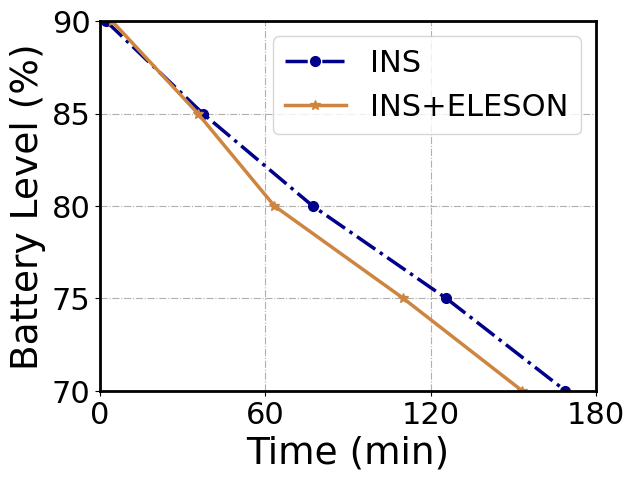}
		\caption{Study on power consumption. }
		\label{fig: exp_power}
	\end{minipage}
\end{figure*}

Figure~\ref{fig: exp_crowd} shows the performance of \sysname{} on unseen elevators and escalators. 
In this experiment, we separate training and testing data by shopping malls, such that the conveyors in testing are unseen to the model. 
In the figure, the F1 score of \sysname{} slightly reduces by around 3\% compared with previous results, due to the subtle shifts of motion patterns and magnetic environments. Despite so, with the enhanced robustness, \sysname{} achieves satisfactory F1 scores (more than 0.85) on unseen elevators and escalators. This validates the generality of \sysname{} in practice. 

We further evaluate \sysname{} beyond mall scenarios through a crowdsourcing experiment, and the results are shown in Figure~\ref{fig: exp_crowdsource}. 
In the experiment, we collect INS signals from users' daily usage and let the users to mark the time slots (by 15{\em min}) in which they used a conveyor. Also, the data are from the user devices that are unseen in the training data. We regard it as a true positive if \sysname{} recognizes the conveyor in a marked slot or if it remains ``neither'' state in an unmarked slot. From the figure, \sysname{} performs similarly to the experimental results in shopping malls, yet the F1 score may slightly slide due to the noisy labels. This validates the generality of \sysname{} in more real-world scenarios.    

Figure~\ref{fig: exp_roc} shows the comparison of different confidence estimation approaches on conveyor state classification by ROC curves. 
The curve indicates a better discriminability of confidence when it is closer to the upper-left corner, and vice versa.  
In the figure, the entropy-based approach (Softmax) shows poor discriminability due to overconfidence. 
TempScale universally reduces Softmax confidence, but its improvement in discriminability is limited. 
While the Bayesian approach can capture epistemic uncertainty, its performance is not stable due to the sampling nature, and more sampling operations lead to heavier computations that are not favorable for the mobile computing.  
Overall, the evidential state classifier shows strong discriminability of confidence of 0.81 in AUROC with lightweight computation. 

Figure~\ref{fig: exp_add_dirich} shows the ablation study on the uncertainty optimization in Equation~(\ref{eq:dirich_var}). The classifier without uncertainty optimization is labeled by ``Without''. In the figure, the uncertainty optimization does not influence the F1 score, because it optimizes the evidence collection over all states. This validates that the uncertainty optimization can improve the confidence discriminability without compromising classification effectiveness. 

\subsection{System Parameters}
\label{subsec: result}

In this section, we study the system parameter of \sysname{} with the confidence threshold of 0.  

Figure~\ref{fig: exp_inv} shows how the F1 score varies with the weight $w_1$ in the loss function of causal feature extractor in Equation~(\ref{eq: cf loss}). 
From the figure, the F1 score increases with the weight when it is less than 0.4 and flats off after that. 
The gain is because the reconstruction prevents the information loss from decomposition. 
In the experiment, we use $w_1=0.6$. 

Figure~\ref{fig: exp_consist} plots how the F1 score varies with the weight $w_2$ in the loss function of causal feature extractor in Equation~(\ref{eq: cf loss}).
In the figure, the accuracy shows a U-shape as the weight increases. 
The F1 score increases because the constraint stabilizes the causal feature. 
The decrease, on the other hand, is when the weight is so large that the extracted features become inflexible.  
In the experiment, we use $w_2=0.3$. 

Figure~\ref{fig: exp_beh} shows how the F1 score varies with the weight $w_3$ in the loss function for the magnetic feature extractor in Equation~(\ref{eq: overall_loss}).
Similar to Figure~\ref{fig: exp_consist}, the F1 score shows a U-shape varying with the weight. 
The increases because the behavior filter enhances the differential feature to be robust against pedestrian behaviors.  
However, leaning too much on adversarial learning may cause the feature to be inflexible. 
In the experiment, we use $w_3=0.4$. 

In Figure~\ref{fig: exp_mag}, we compare the different implementations for the differential feature extractor in Equation~(\ref{eq: variation_feature}). 
Specifically, ``Raw3D'' stands for the 3D magnetic signal, ``Raw1D'' is the intensity of the magnetic signals, ``Diff3D'' is the temporal differential of the magnetic signals, ``Diff1D'' is the differential of the intensity, and ``VarHist'' is the histogram on ``Diff1D''. 
In the figure, the differential features generally achieve higher F1 score than the raw signals because they are more independent of locations. On the other hand, the intensity of magnetic signals outperforms the 3D orientation as it reduces noise. Finally, the ``DiffHist'' fails to improve the F1 score as it reduces the important temporal feature of conveyor states. 

In Figure~\ref{fig: exp_w4}, we study how the F1 score varies with the weight $w_4$ in Equation~(\ref{eq: overall_loss}), which is the weight for the loss function of the causal feature extractor. In the figure, the F1 score increases with the weight because the loss function supervises the extraction of the conveyor causal features. In the experiment, we use $w_4=1$, where the F1 score flattens off after that. 

Figure~\ref{fig: exp_ud_conf} shows the distribution of confidence over the whole dataset. The figure shows that the UD ratio grows exponentially with the confidence threshold, with a short tail on the left side. This indicates that \sysname{} is confident about most decisions. 
In the experiment, we set the confidence threshold as 0.5, which leads to a UD ratio of 5\%. 

\subsection{Efficiency Study}
\label{subsec: exp_on_device}
In this section, the efficiency studies are conducted on a mobile phone for illustrative purposes, and the conclusion is not limited to the experimental device.  

Figure~\ref{fig: exp_edge} shows how the error rate and inference time (for making one prediction) vary with the model size. 
The model size is adjusted by the neuron quantity of each layer, and the error rate equals one minus the F1 score.  
In the figure, the inference time increases with model size, while the error kneels down at the model size of 9MB. 
Therefore, we choose the model size of 9MB, which is around 0.3\% of the phone memory, and each model inference takes around 0.4 seconds. 
With this setting, \sysname{} operates in real time (i.e., inference time is less than step size) with a minimal memory budget. 

Finally, we show the power consumption of \sysname{} on a smartphone in Figure~\ref{fig: exp_power}. 
In the experiment, we turn on the INS sensor and use \sysname{} to process the INS signals. 
In contrast, we maintain the phone states and turn on the INS sensor without running \sysname{}. 
From the figure, 
INS consumes around 18\% of the battery (from 90\% to 72\%) in 150 minutes, and running \sysname{} only additionally takes 2\% more battery in the same duration.  
This validates the minimal cost of \sysname{} in terms of power consumption.

\section{Conclusion}\label{sec: conclude}

In this paper, we study classifying the conveyor state of a pedestrian with arbitrary behaviors to elevator, escalator, or neither, or simply the conveyor state classification, using the inertial navigation system (INS) on his/her smartphone (i.e., accelerometer, gyroscope, and magnetometer). This research problem is fundamental to many smart city applications, such as indoor navigation and people flow management. The challenge is posed by the arbitrary behaviors of pedestrians, because they entangle with the conveyor states, perturb INS signals, and obscure the classification decision on the states. 

We propose \sysname{}, a novel, effective, and lightweight deep learning approach that classifies the conveyor state under arbitrary pedestrian behaviors using phone INS without the need for any behavior labeling. \sysname{} separates the motion features of moving elevators and escalators from pedestrian behaviors based on causal decomposition and extracts the magnetic feature of conveyor states based on adversarial learning. Given those features, it uses an evidential classifier to estimate the confidence of each state, which reflects the similarity of an input INS signal to its training data. Through extensive experiments on 36,420 instances of conveyor state data with arbitrary unlabeled pedestrian behaviors collected from ten shopping malls, \sysname{} shows satisfactory performance, achieving high accuracy of over 0.9 in F1 score and sound confidence discriminability of 0.81 in AUROC (Area Under the Receiver Operating Characteristics), which improves from previous approaches by 14\% in F1 score. Additionally, our efficiency study demonstrates \sysname{} to operate on a mobile phone in real time (0.4s for one inference), requiring only 9MB memory usage and consuming merely 2\% battery in 2.5 hours. 

\sysname{} is a pioneering work on classifying the conveyor states of pedestrians using deep learning. 
In the future, we will extend the scheme to infer the direction of transport or floor transition, and incorporate barometer reading to accomplish richer tasks, higher accuracy, and stronger robustness.
We would also like to cover other conveyor types, such as travelators and wheelchairs.   

\section{Acknowledgment}
We would like to thank HONOR for introducing the problem to us and sharing the data for our study.

\bibliographystyle{IEEEtran}
\bibliography{reference}

\label{bio}

\begin{IEEEbiography}[{\includegraphics[width=1in,height=1.25in,clip,keepaspectratio]{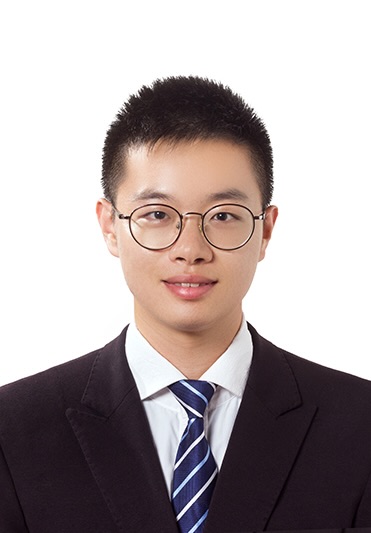}}]{Tianlang He} received his Bachelor of Engineering degree (with honor) from Donghua University, Shanghai, China, in 2018. He obtained his Master of Science degree from the Hong Kong University of Science and Technology (HKUST), Hong Kong, China, in 2019. Currently, he is a Ph.D. candidate in the Department of Computer Science and Engineering at HKUST. His research interests include AI Internet of Things (AIoT) and cyber-physical system (CPS), with a focus on system robustness. 
\end{IEEEbiography}

\begin{IEEEbiography}[{\includegraphics[width=1in,height=1.25in,clip,keepaspectratio]{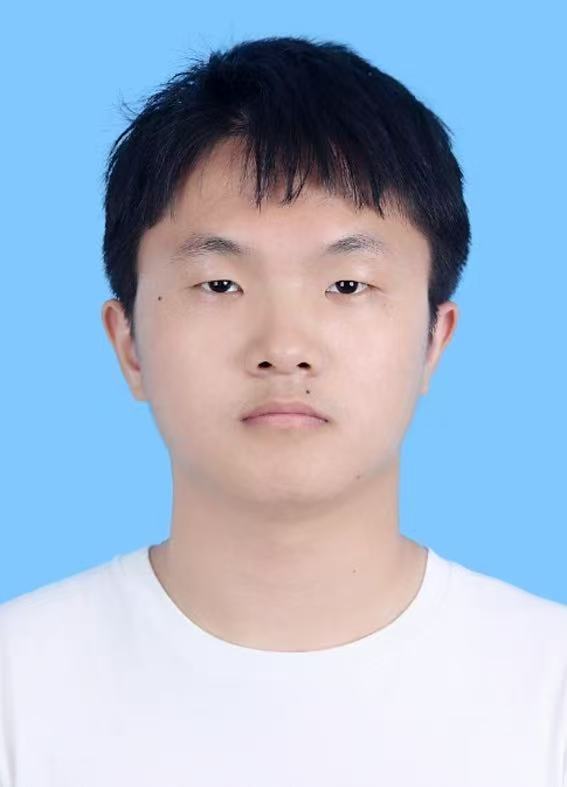}}]{Zhiqiu Xia}
received the bachelor of science degree (with honor) in data science and technology from the Hong Kong University of Science and Technology (HKUST), Hong Kong, China, in 2024. He is currently working toward the PhD degree in electrical and computer engineering at Rutgers University, NJ, USA. His research interest includes machine learning and large language model.
\end{IEEEbiography}

\begin{IEEEbiography}[{\includegraphics[width=1in,height=1.25in,clip,keepaspectratio]{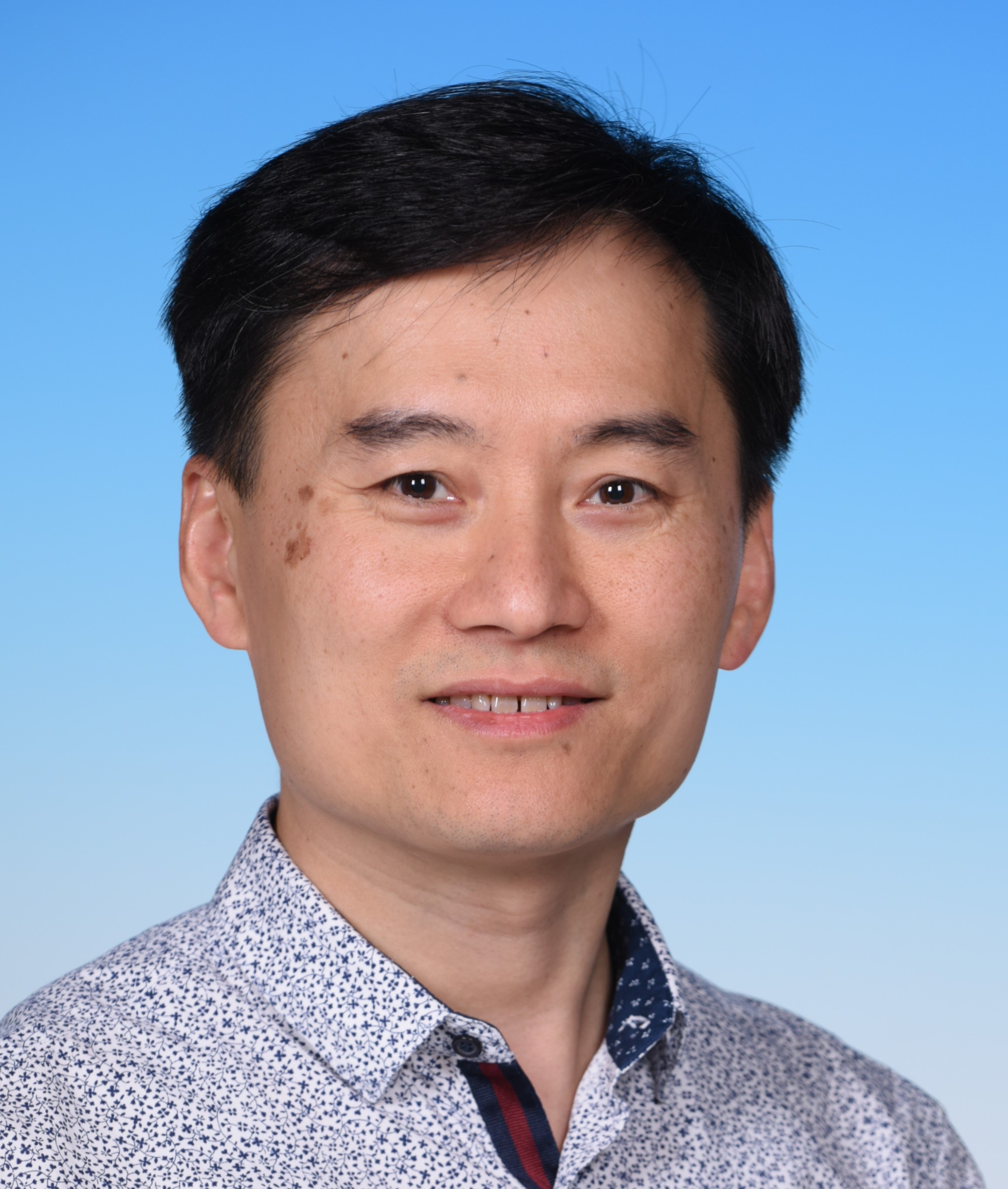}}]{Dr. S.-H. Gary Chan} is currently Professor of the Department of Computer Science and Engineering and Associate Director of GREAT Smart Cities Institute, The Hong Kong University of Science and Technology (HKUST), Hong Kong.  He is also Board Director of Hong Kong Logistics and Supply Chain MultiTech R\&D Center (LSCM).  He received MSE and PhD degrees in Electrical Engineering with a Minor in Business Administration from Stanford University (Stanford, CA).  He obtained his B.S.E. degree (Highest Honor) in Electrical Engineering from Princeton University (Princeton, NJ), with certificates in Applied and Computational Mathematics, Engineering Physics, and Engineering and Management Systems.  His research interests include smart IoT and sensing systems, edge AI, location AI and mobile computing, video/user/data analytics, technology transfer and entrepreneurship.

Professor Chan has been Vice-Chair of Peer-to-Peer Networking and Communications Technical Sub-Committee of IEEE Comsoc Emerging Technologies Committee, steering committee member and TPC chair of IEEE Consumer Communications and Networking Conference (IEEE CCNC), and area chair of the multimedia symposium of IEEE Globecom and IEEE ICC.  He has been Associate Editor of IEEE Transactions on Multimedia, Guest Editor of ACM Transactions on Multimedia Computing, Communications and Applications, IEEE Transactions on Multimedia, IEEE Signal Processing Magazine, IEEE Communication Magazine, etc.

Through technology transfer and entrepreneurship, Professor Chan has successfully deployed his research results in industry and co-founded several startups with high commercial and societal impacts.  His innovations have received numerous awards and recognitions over the years.  Notably, he received Hong Kong Chief Executive's Commendation for Community Service for "outstanding contribution to the fight against COVID-19".  He was the recipient of Google Mobile 2014 Award
and Silver Award of Boeing Research and Technology.  He was a visiting professor or researcher in Microsoft Research, Princeton University, Stanford University, and University of California at Davis.  At HKUST, he was Director of Entrepreneurship Center, Director of Sino Software Research Institute, Co-director of Risk Management and Business Intelligence program, and Director of Computer Engineering Program. He was a William and Leila Fellow at Stanford University, and the
recipient of the Charles Ira Young Memorial Tablet and Medal and the POEM Newport Award of Excellence at Princeton University.  He is elected Fellow of Sigma Xi (FSX) and Chartered Fellow of The Chartered Institute of Logistics and Transport (FCILT).

\end{IEEEbiography}

\end{document}